\pdfoutput=1
\documentclass[10pt,twocolumn,letterpaper]{article}

\usepackage[pagenumbers]{cvpr}

\usepackage{booktabs,multirow,graphicx,xcolor,pifont}
\newcommand{\up}{$\uparrow$}
\newcommand{\down}{$\downarrow$}

\definecolor{cvprblue}{rgb}{0.21,0.49,0.74}
\usepackage[pagebackref,breaklinks,colorlinks,allcolors=cvprblue]{hyperref}
\def\confName{CVPR}
\def\confYear{2027}

\usepackage{xspace}
\def\model{MultiControl\xspace}

\title{\model: A Geometry-Aware Multi-Agent Egocentric Video World Model with 3D Full-Body Pose Control}

\author{
{\fontsize{11}{13}\selectfont Sina Mokhtarzadeh Azar\textsuperscript{*1,2} \quad
Enrico Pallotta\textsuperscript{*1,2} \quad
Sina Raoufi\textsuperscript{1} \quad
Lars Doorenbos\textsuperscript{1,2} \quad
Serdar Ozsoy\textsuperscript{1,2}} \\[0.1cm]
{\fontsize{11}{13}\selectfont Umar Iqbal\textsuperscript{3} \quad
Juergen Gall\textsuperscript{1,2}} \\[0.1cm]
{\normalsize \textsuperscript{1}University of Bonn \quad
\textsuperscript{2}Lamarr Institute for Machine Learning and Artificial Intelligence \quad
\textsuperscript{3}NVIDIA} \\
{\footnotesize \textsuperscript{*}Equal Contribution}
\\[0.1cm]
{\normalsize \href{https://cvg-bonn.github.io/MultiControl/}{cvg-bonn.github.io/MultiControl}}
}

\begin{document}
\twocolumn[{%
\renewcommand\twocolumn[1][]{#1}%
\maketitle
\vspace{-1.2cm}
\begin{center}
    \centering
    \captionsetup{type=figure}
    \includegraphics[width=0.9\textwidth]{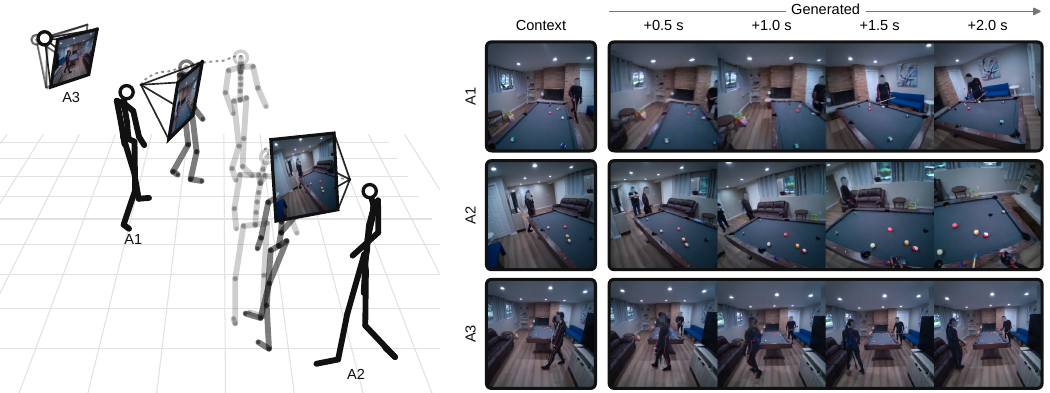}
    \captionof{figure}{\model generates egocentric videos of several agents in one scene. Each agent follows its own
  3D motion (left, future poses faded) with flexible control: full body for A1 and A2, head only for A3. All views are generated jointly and show the same world (right).}
    \label{fig:teaser}
\end{center}%
 }]
\begin{abstract}
Video world models are capable of generating and exploring new worlds mostly through the perspective of a single agent. These isolated worlds do not take into account the other agents that may share the same world and interact with it and with one another.
We introduce \textbf{\model}, the first multi-agent egocentric video world model with 3D full-body pose control, as a step toward shared worlds where people can act and interact.
To keep the views of agents mutually consistent, we introduce cross-view attention blocks with URoPE as the geometric positional embedding. We also propose Body-Ray, a second branch that conditions each view on the 3D joints of the other agents. We design a training recipe that makes \model flexible. The same model handles a variable number of agents, takes any available context views as input, and works whether or not an agent's full-body pose is given.
Finally, we propose an evaluation based on counterfactual edits of the body poses, which tests whether the generated views follow changes in each agent's motion. \model outperforms all compared multi-agent baselines in visual quality and in reproducing the poses of other agents across views.

\end{abstract}

\section{Introduction}
\label{sec:intro}

People rarely act alone. Instead, they act together in a shared environment: they walk side by side, hand objects to each other, or talk face to face, and observe each other while doing so.
Egocentric world models, which generate first-person video from the actor's own body motion~\cite{bai2026whole, tu2026playerone, pallotta2026egocontrol}, are
a promising step toward simulating such behavior for virtual reality and embodied agents, but they can only simulate a
single person.
Recently, multi-agent video world models have started to address this setting~\cite{savva2026solaris, wu2026multiworld,
liu2026gamma, hu2026multiplayer, zhu2026shareverse}: they generate one video
per agent and couple the videos so that all agents live in the same world.
However, these models are trained and tested mostly in games such as Minecraft, or with fixed tabletop cameras, and they are controlled by relatively simple action spaces, with agents represented by mostly rigid avatars seen through an almost perfect, undistorted view.
While some works now move to more realistic scenes~\cite{sun2026prisma, hu2026metaworld}, their agents are controlled through camera motion only.
As a result, multi-agent world modeling of real human interaction, where each person acts with a full body, remains, to our knowledge, unexplored.

Creating a multi-agent world model is challenging due to several factors, and we argue that a successful multi-agent world model must satisfy three requirements, on top of producing realistic video.
\textbf{Per-agent control}: each agent's camera and body follow its own pose in its own view.
\textbf{Cross-agent action coherence}: the motion generated for one agent appears, in sync and at the right
place in 3D, in the view of every other agent that sees it.
Existing benchmarks only coarsely test for this, by asking vision-language models whether the other agent is visible or moved left or right~\cite{savva2026solaris}.
We therefore introduce a counterfactual-edit protocol, where we replace the real motion of the agents with a different one, estimate the 3D pose of each edited agent in the views of the others, and compare it with the control pose.
\textbf{Shared-world consistency}: all views show the same room and the same agents, including views generated
without any context frames.
This is the standard multi-view requirement, made far more challenging in the multi-agent egocentric setting by fisheye lenses, cameras that move every frame, and real people.

To address these challenges, we present \textbf{\model}, the first multi-agent egocentric video world model in which each agent is controlled by a 3D full-body pose.
Starting from a single-agent model, \model generates the views of all
agents jointly, which we couple by introducing a cross-view attention layer in every block.
We train \model on the video recordings of Nymeria~\cite{ma2024nymeria}, in which every agent wears synchronized
head-mounted cameras.
Our cross-view attention of \model has a zero-initialized output, enabling training to start from a single-agent model such as EgoControl~\cite{pallotta2026egocontrol}.
For shared-world consistency, each token must know where it lies in the other views.
We adapt the geometric position encoding of URoPE~\cite{xie2026urope} to fisheye cameras that move every frame.
For cross-agent action coherence, an optional Body-Ray branch gives each view the 3D joints of the
other agents.
We also make use of a flexible training recipe, where any subset of views may receive context frames and an agent may also be
controlled by its head motion only.
By doing so, the same model can control an agent without a recorded body pose like any other, or generate the views of
two agents entirely from the context frames of a third agent.

\noindent In summary, our contributions are:
\begin{itemize}
  \item We present \model, the first multi-agent egocentric video world model controlled by 3D full-body poses of multiple agents, trained on real videos.
  \item We introduce cross-view attention with geometric position encoding for consistency across the views of agents, and Body-Ray to
  explicitly consider the 3D joints of the other agents in each view.
  \item We use a flexible training recipe that randomly drops body poses and context frames, enabling capabilities such as
  generating multiple synchronized egocentric views from a single exocentric view.
  \item With a focus on counterfactual edits of the body poses, we introduce an evaluation protocol for cross-agent action coherence and shared-world consistency.

\end{itemize}

\section{Related Works}
\label{sec:relatedworks}
\noindent
\textbf{Single-agent world models}
Egocentric videos offer a privileged perspective for learning from human behavior, as they capture embodied demonstrations directly from the actor's point of view~\cite{grauman2022ego4d, grauman2024ego}. In parallel, video world models have proven to be valuable simulators~\cite{bar2025navigation, agarwal2026cosmos} also thanks to foundational works that enabled control over the egocentric viewpoint through camera motion~\cite{he2025cameractrl, wang2024motionctrl}. More recently, egocentric world models have moved towards richer forms of interaction, following two main lines of work. The first focuses on dexterity, conditioning generation on hand poses to model fine-grained hand-object interactions~\cite{Kim_2026_CVPR, li2026egocentric}. The second targets full-body pose control for immersive navigation and interaction~\cite{bai2026whole, tu2026playerone, pallotta2026egocontrol}. However, these approaches model a single egocentric agent, whereas human activities often involve multiple agents interacting in a shared environment. This direction remains underexplored, and it is the one we address in this work.

\noindent
\textbf{Multi-agent world models}
Solaris~\cite{savva2026solaris} pioneered two-agent generation in Minecraft by extending self-attention across both views and adding fixed player-ID embeddings.
Subsequent works addressed problems specific to multi-agent world models: MultiWorld~\cite{wu2026multiworld} introduced a global state encoder built on VGGT~\cite{Wang_2025_CVPR} features, while Gamma-World~\cite{liu2026gamma} proposed sparse-hub attention avoiding self-attention cost to grow quadratically with the number of agents.
These methods target games or fixed tabletop robot cameras, and thus do not face the challenges posed by real egocentric human videos.
A parallel line of work targets realism and cross-view consistency~\cite{sun2026prisma, hu2026metaworld, zhu2026shareverse}, but restricts action to simple camera motion.
Although these methods rely on realistic synthetic data, they do not address egocentric data, with its complex dynamics and the need to model the articulated motion of multiple agents sharing the same environment.

\noindent
\textbf{Geometric position encodings}
A transformer attending across camera views needs to know how they relate in 3D.
Cameras can be given as input features, as parameters~\cite{wang2024motionctrl} or Pl\"ucker ray maps~\cite{he2025cameractrl}, leaving the network to learn cross-view correspondences by itself.
Other approaches encode the relative geometry directly in the attention.
GTA~\cite{miyato2024gta} and CaPE~\cite{Kong_2024_CVPR} make the attention between two tokens depend on the relative camera poses. PRoPE~\cite{prope}, extends this to the relative projective transformation between the two cameras including their intrinsics, Prisma-World~\cite{sun2026prisma} uses it to modulate cross-agent attention.
The most recent encodings describe each token instead of each camera pair.
RayRoPE~\cite{wu2026rayrope} places every patch at a predicted depth along its ray and encodes this point in the coordinates of the query camera. URoPE~\cite{xie2026urope} lifts every key patch to fixed depths along its ray, projects these points into the query image, and applies RoPE~\cite{su2024roformer} at the projected positions.
These encodings were mainly studied for novel view synthesis and 3D perception.
We adapt URoPE to relate the egocentric videos of several moving people,  encode the other agents' 3D joints in the same way,
and compare Pl\"ucker ray maps, PRoPE and URoPE in our setting.

\section{Method}
\label{sec:method}

We consider $N$ agents with egocentric views who act in the same environment. Our goal is then to generate the future frames $Y_i$ of each agent $i\in\{1,\dots,N\}$ based on the context frames $C_i$ that each agent has observed up to the present. Each agent is further controlled by two signals: its 3D body pose sequence $P_i$ and its corresponding camera trajectory $T_i$, where the camera poses are represented in a {world coordinate system} that is shared by all agents. A multi-agent world model thus predicts the future views $Y_{1:N}=\{Y_i\}_{i=1}^{N}$ of all agents from their context frames $C_{1:N}$ and controls $P_{1:N}$ and $T_{1:N}$ by estimating the joint distribution
\begin{equation}
p\big(Y_{1:N}\mid C_{1:N},\,P_{1:N},\,T_{1:N}\big).
\label{eq:task}
\end{equation}

This task is challenging since the generated video frames should not only follow the controls of its own agent, but also consider how the other agents act in the environment. In other words, the generated frames of the egocentric views of all agents need to be consistent. This includes the poses of the other agents, objects that have been seen by only some of the agents before, and changes of the environment that are done by one agent, like moving an object. This makes egocentric video generation of multiple agents much more difficult than video generation of a single agent.

\begin{figure*}[t]
\centering
\begin{minipage}[t]{0.63\textwidth}
\vspace{0pt}
\centering
\includegraphics[width=\linewidth]{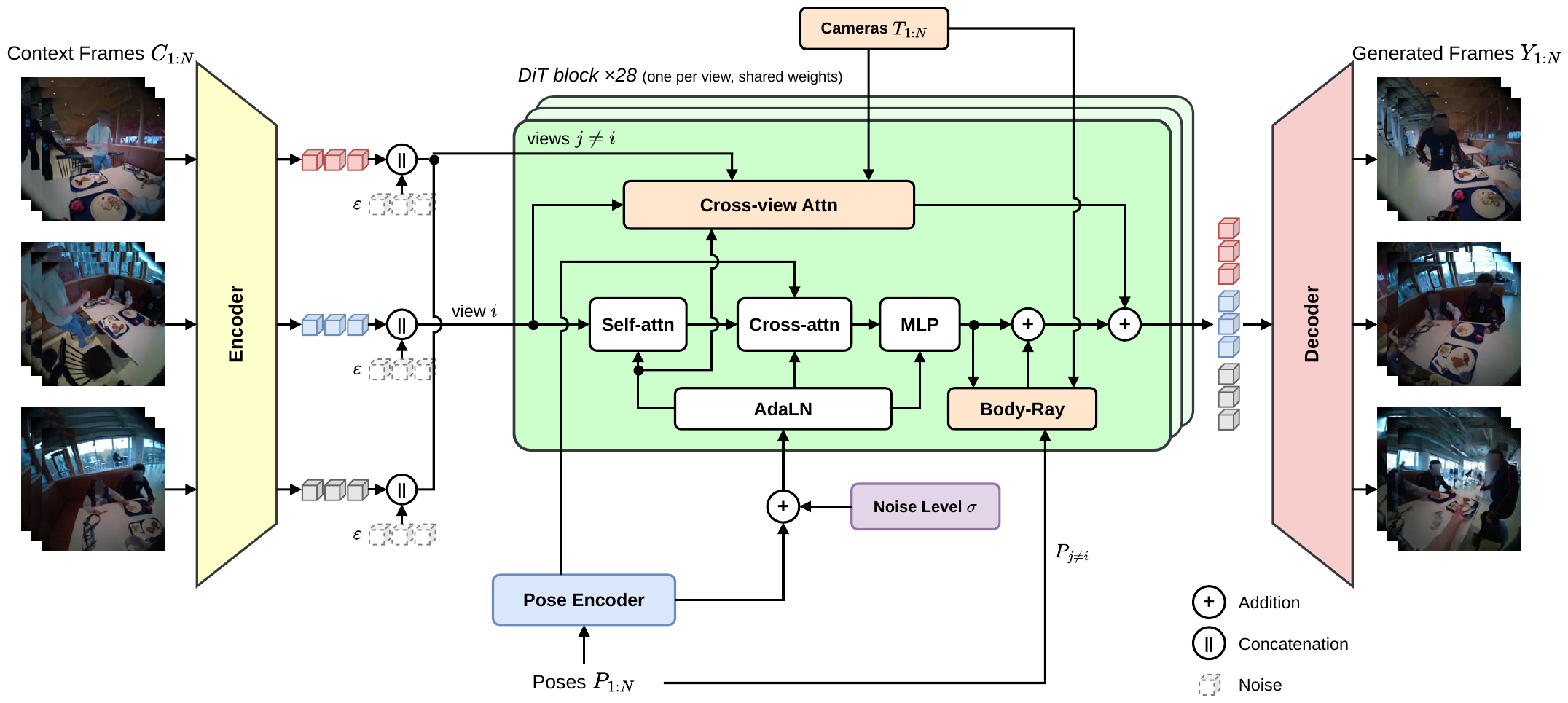}
\end{minipage}\hfill
\begin{minipage}[t]{0.35\textwidth}
\vspace{0pt}
\centering
\includegraphics[width=\linewidth]{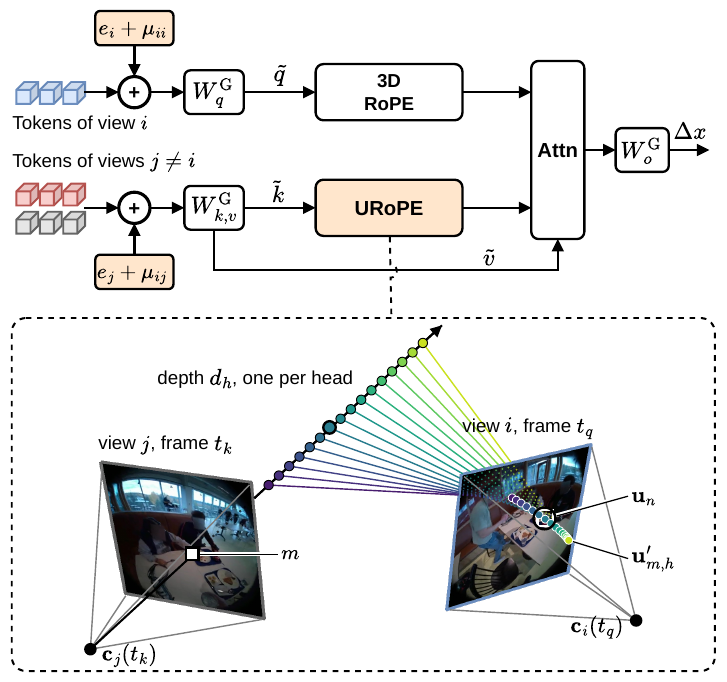}
\end{minipage}
\caption{\textbf{Overview of \model (left) and cross-view attention with geometric position encoding (right).}
\textbf{Left:} The views of all agents are generated jointly by a DiT with shared weights across agents. Each agent is conditioned on its own pose and noise level, while a cross-view attention layer gives access to the tokens of the other agents $j\neq i$, using the camera poses of all agents.
\textbf{Right:} The queries in the cross-view attention layer come from agent $i$ whereas keys and values are from the other agents $j\neq i$. While $e_i$ encodes each agent, The $\mu_{ij}$ encodes the pose of camera $j$ relative to camera $i$. URoPE places each token of view $j$ at fixed depths along its pixel ray and projects these points into view $i$, so that the attention score depends on how close a projected key lies to the query in view $i$.
}
\label{fig:pipeline}
\label{fig:crossview}
\end{figure*}

An overview of our approach is given in \Cref{fig:pipeline}. Our approach builds on a pretrained single-agent model that generates one egocentric video from its wearer's pose (\cref{sec:base}). This model is shared among the agents, and each model takes its own context $C_i$ and 3D body pose sequence $P_i$ as input. The agents exchange information through a cross-view attention layer in every block (\cref{sec:crossview}), which uses the 3D geometry of the cameras (\cref{sec:urope}). The model is trained (\cref{sec:training}) such that it is not limited to a fixed number of agents and that not all inputs $C_{i}$, $P_{i}$, and $T_{i}$ need to be provided for all agents. For example, the context frames of some agents can be omitted, i.e., we generate the view of an agent given only the previous observations of the other agents.

\subsection{Single-Agent Model}
\label{sec:base}

To build a multi-agent world model, we start from a single-agent that generates a future egocentric video of one person from its context frames and the person's 3D body pose, i.e., it models $p(Y_i\mid C_i,P_i)$ for a single agent. In our implementation, we use EgoControl~\cite{pallotta2026egocontrol} since it is publically available. It represents the pose with relative transforms for every frame $t$:
\begin{equation}
\Delta H_t = H_t\,H_{t-1}^{-1},
\label{eq:relpose}
\end{equation}
where $H_t\in SE(3)$ is the pose of the head at frame $t$. The motion of the pelvis is computed in the same way, and the other 21 joints are represented relative to the pelvis.

EgoControl is a latent video diffusion model.
Its diffusion transformer $D_\theta$ denoises the video latents
$z_\sigma = z_0 + \sigma\varepsilon$ at noise level $\sigma$,
while the context latents are kept noise-free.
It is trained with the weighted denoising loss~\cite{karras2024edm2}:
\begin{equation}
\mathcal{L}=\mathbb{E}_{z_0,\sigma,\varepsilon}\big[\,w(\sigma)\,\lVert D_\theta(z_\sigma;\sigma,C_i,P_i)-z_0\rVert_2^2\,\big],
\label{eq:loss}
\end{equation}
where $w(\sigma)$ weights each noise level.
The pose conditions every block in two ways.
First, an embedding of the pose sequence is added to the noise-level embedding.
It determines the scale $\gamma$, the shift $\beta$ and an output gate
of the adaptive layer normalization in each block,
\begin{equation}
\mathrm{AdaLN}(x)=\mathrm{LN}(x)\odot(1+\gamma)+\beta,
\label{eq:adaln}
\end{equation}
where $x$ is the hidden state of a token and $\mathrm{LN}$ denotes layer normalization.
Second, the video tokens attend through cross-attention to the pose tokens,
one per frame.

\subsection{Multi-Agents with Cross-view Attention}
\label{sec:crossview}

For multi-agent video generation with $N$ agents, we use the base model (\Cref{sec:base}) shared across all agents with the same weights. However, we need to make substantial changes to model the joint distribution of all agents (\cref{eq:task}). To this end, we propose a cross-view attention layer and add it to every block. It ensures that scene information visible by one agent is shared with the other agents. This also includes objects that were not visible in the context frames of any agent since the generation of unseen parts of the environment also needs to be consistent among agents. The layer utilizes the camera trajectories $T_i$ of each agent which define the geometry between the views.

\noindent
\textbf{Cross-view attention.}
In every view, the layer takes the same input as the block's self-attention:
the normalized hidden state $\hat{x}_n=\mathrm{AdaLN}(x_n)$ of each token $n$
(\cref{eq:adaln}). A token $n$ of agent $i$ at latent frame $t_q$ queries the tokens of all other agents $j\neq i$ at all of their latent frames, including the context frames. We denote these key tokens by $\mathcal{K}(i)$. Each key token $m \in \mathcal{K}(i)$ belongs to
an agent $j \neq i$ and a latent frame $t_k$.
With the attention head index omitted, the query of token $n$ and the key and value of token $m$ are
\begin{equation}
\begin{aligned}
\tilde{q}_n &= \mathrm{Norm}_q\big(W^{\mathrm{G}}_q(\hat{x}_n + e_i + \mu_{ii})\big),\\
\tilde{k}_m &= \mathrm{Norm}_k\big(W^{\mathrm{G}}_k(\hat{x}_m + e_j + \mu_{ij}(t_k))\big),\\
\tilde{v}_m &= W^{\mathrm{G}}_v\big(\hat{x}_m + e_j + \mu_{ij}(t_k)\big),
\end{aligned}
\label{eq:qkv}
\end{equation}
where $\mathrm{Norm}_q$ and $\mathrm{Norm}_k$ are per-head RMS normalizations. The learned codes $e_i$ and $\mu_{ij}$ provide information about the agent $i$ and the relative position of agent $j$ with respect to agent $i$. These codes are important to obtain consistently generated video frames across agents. While $\tilde{q}_n$, $\tilde{k}_m$, and $\tilde{v}_m$ will be used to compute the cross-attention, see \Cref{fig:pipeline} (right), we now describe how $e_i$ and $\mu_{ij}$ are obtained.

The ID embedding $e_i$ tells the agents apart and is constant over $t$. We learn a small pool of embeddings and randomly sample without replacement an embedding per agent, so that no embedding is tied to a particular agent. Therefore, the model can also generate more agents at test time (see supplementary).
To obtain the camera embedding $\mu_{ij}(t)$, we first map the pose of camera $j$ into the coordinates of camera $i$ at frame $t$ by $T_i(t)^{-1}T_j(t)$.  For a relative camera pose $T$ with rotation $R$ and translation $\mathbf{c}$, we compute the vector $\phi(T)=\big(\mathbf{c},\ \log(1+\lVert\mathbf{c}\rVert),\ R^{(1)},\ R^{(2)}\big)$, where $R^{(1)}$ and $R^{(2)}$ are the first two columns of $R$. We finally apply an MLP $\psi$ to obtain $\mu_{ij}(t)$:
\begin{equation}
\mu_{ij}(t)=\psi\big(\phi\big(T_i(t)^{-1}\,T_j(t)\big)\big).
\label{eq:camcode}
\end{equation}
The query uses $\mu_{ii} = \psi(\phi(I))$, where $I$ is the identity matrix.

Using $e_i$ and $\mu_{ij}$, the queries $\tilde{q}_n$, keys $\tilde{k}_m$, and values $\tilde{v}_m$ are computed as in \cref{eq:qkv} for each token $n$ and $m$. The cross-attention is then computed by
\begin{equation}
\Delta x_n = W^{\mathrm{G}}_o \sum_{m\in\mathcal{K}(i)} \operatorname{softmax}_m\big(s(n,m)\big)\,\tilde{v}_m,
\label{eq:crossview}
\end{equation}
where $s(n,m)=\tilde{q}_n^\top\tilde{k}_m/\sqrt{D}$, with $D$ being the dimension of an attention head. The feature of token $n$ of agent $i$ is then updated by $x_n\leftarrow x_n+\Delta x_n$. While this approach already considers the relative positions of the agents' cameras, it does not consider where each token will be visible in the views of the other agents. We therefore modify the cross-attention matrix $s(n,m)$ as we explain next.

\subsection{Geometric Position Encoding}
\label{sec:urope}

For a single agent, we can encode the position of a token at latent frame $t$ and grid position $\mathbf{u}$ (row and column) with 3D RoPE~\cite{su2024roformer}, which we denote by $\mathcal{M}(t,\mathbf{u})$. Since $\mathcal{M}(t_1,\mathbf{u}_1)^\top\mathcal{M}(t_2,\mathbf{u}_2)=\mathcal{M}(t_2-t_1,\mathbf{u}_2-\mathbf{u}_1)$, the relative position of two tokens is given by their difference.

\noindent Across views, this difference is not meaningful because the same grid position shows different parts of the scene in different views. To encode whether the key token $m$ belongs to the same part of the scene as a given query token $n$, we encode the position of a key relative to the query through the 3D geometry of the cameras, inspired by URoPE~\cite{xie2026urope}.

Let key token $m$ belong to view $j$ at latent frame $t_k$, and let $\mathbf{r}_m$
be the direction of its pixel ray in the coordinates of camera $j$, scaled so that
its forward component is 1.
We obtain $\mathbf{r}_m$ by unprojecting the token centre with the calibrated
camera model.
Each attention head $h$ places the key at a fixed depth $d_h$ along this ray.
It then projects the resulting 3D point into the camera view $i$
at the query's latent frame $t_q$:
\begin{equation}
\begin{aligned}
\mathbf{X}_{m,h} &= \mathbf{c}_j(t_k) + d_h\,R_j(t_k)\,\mathbf{r}_m,\\
\mathbf{u}'_{m,h} &= \pi\Big(R_i(t_q)^\top\big(\mathbf{X}_{m,h}-\mathbf{c}_i(t_q)\big)\Big),
\end{aligned}
\label{eq:lift}
\end{equation}
where $\pi$ is the projection of the calibrated camera model, expressed in units
of the token grid.
The score of head $h$ then compares the query's own position with the key's
projected position:
\begin{equation}
s_h(n,m)=\frac{1}{\sqrt{D}}\,\tilde{q}_{n,h}^\top\,
\mathcal{M}\big(t_k-t_q,\ \mathbf{u}'_{m,h}-\mathbf{u}_n\big)\,\tilde{k}_{m,h}.
\label{eq:score}
\end{equation}
In practice, the query is rotated by the RoPE encoding $\mathcal{M}(t_q,\mathbf{u}_n)$, and the
key by $\mathcal{M}(t_k,\mathbf{u}'_{m,h})$, that is, at its projected position
instead of its own.
The values are not rotated.

When a key token and the query see the same point of the scene, and this point lies at depth $d_h$ in front of camera
$j$, the key's projection $\mathbf{u}'_{m,h}$ lands on the query's position $\mathbf{u}_n$. The attention head $h$ thus treats the
two tokens as if they were at the same place in one image.
As $d_h$ varies, $\mathbf{u}'_{m,h}$ moves along the epipolar curve of the key in the query's image.
Each attention head uses its own depth, spaced logarithmically from close range to a maximum depth, so together the heads
let a query find the tokens of the other views that show the same point of the scene, without predicting depth.
Instead of considering only the relative position of the two cameras, the encoding thus tells each head where every key would
appear in the query's image.

\subsection{Body-Ray}
\label{sec:bodyray}

\begin{figure}[t]
\centering
\includegraphics[width=0.85\columnwidth]{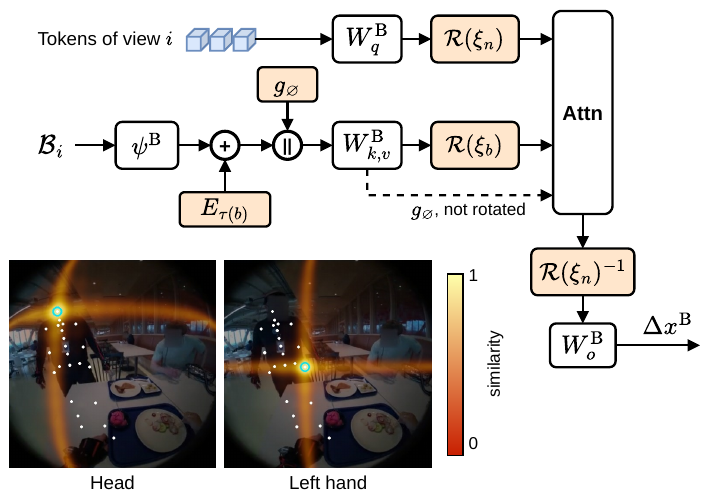}
\caption{\textbf{Body-Ray.}
The image tokens of view $i$ attend to the other people's joints in camera $i$, with queries rotated by the ray angles
$\xi_n$ of the tokens and keys and values by the direction angles $\xi_b$ of the joints (\cref{eq:bodykey,eq:bodyray}).
The panels show, on a real frame, the rotary part of the score between one joint (green circle) and every image token.
}
\label{fig:bodyray}
\end{figure}

Cross-view attention transfers whatever the other views show, the people included, but it has to find out from the
images where things are.
We do not know the 3D layout of the scene, but we do know where the people are, since their 3D poses are given as
controls.
We therefore also propose Body-Ray, an additional branch that conditions each view explicitly on where the other
people's joints are (\cref{fig:bodyray}). We will show that this additional branch will improve the coherence of the generated views.

For view $i$ at latent frame $t$, let $\mathcal{B}_i$ be the joints of the other agents' poses $P_j$, $j\neq i$.
We express the world position $\mathbf{J}_b(t)$ of each joint $b$ in the coordinates of camera $i$,
$\mathbf{p}_b(t)=T_i(t)^{-1}\mathbf{J}_b(t)$.
The key and value of a joint come from a learned embedding $E_{\tau(b)}$ of its joint type $\tau(b)$ and an MLP
$\psi^{\mathrm{B}}$ applied to its log distance and its motion in view $i$,
\begin{equation}
g_b = E_{\tau(b)} + \psi^{\mathrm{B}}\big(\log\lVert\mathbf{p}_b(t)\rVert,\ \mathbf{p}_b(t)-\mathbf{p}_b(t-1)\big).
\label{eq:bodykey}
\end{equation}
Since not all tokens are related to joints, we add a learned null token $g_\varnothing$.

We use a rotary embedding as before using the horizontal and vertical angle $\xi=(\xi^x,\xi^y)=(\operatorname{atan2}(x,z),\operatorname{atan2}(y,z))$ of a direction $(x,y,z)$ in camera coordinates. We write $\xi_n$ for the pixel ray of image token $n$, and $\xi_b$ for the direction from camera $i$ to joint $b$.
The image tokens of view $i$ then attend to these joints and the null entry by
\begin{equation}
\begin{gathered}
s^{\mathrm{B}}(n,b)=\frac{1}{\sqrt{D}}\,q_n^\top\,\mathcal{R}(\xi_b-\xi_n)\,k_b,\\
\Delta x^{\mathrm{B}}_n=W^{\mathrm{B}}_o\sum_{b}\operatorname{softmax}_b\big(s^{\mathrm{B}}(n,b)\big)\mathcal{R}(\xi_b-\xi_n)v_b,
\end{gathered}
\label{eq:bodyray}
\end{equation}
where $q_n=W^{\mathrm{B}}_q\,\mathrm{LN}(x_n)$, $k_b=W^{\mathrm{B}}_k\,\mathrm{LN}(g_b)$. and
$v_b=W^{\mathrm{B}}_v\,\mathrm{LN}(g_b)$.
We use $\xi_\varnothing=0$ for the null entry.
The score of a joint thus depends on the positions only through the angle difference $\xi_b-\xi_n$, that is, on where
the joint lies relative to the token's line of sight.
As in~\cite{miyato2024gta}, the values are rotated as well, so the output also carries the direction of each
attended joint relative to the token's line of sight.
In practice, since $\mathcal{R}(\xi_n)^\top\mathcal{R}(\xi_b)=\mathcal{R}(\xi_b-\xi_n)$, the query is rotated by
$\mathcal{R}(\xi_n)$, the keys and values of the joints by $\mathcal{R}(\xi_b)$, and the output back by
$\mathcal{R}(\xi_n)^{-1}$.
The branch is added to every block, after the block and before the cross-view update, with $W^{\mathrm{B}}_o$
initialized to zero so that training starts from the model without Body-Ray.

\subsection{Training}
\label{sec:training}

All views of a sample are denoised jointly with the loss of \cref{eq:loss}, each view at its own noise level
$\sigma_i$.
To obtain one model for many combinations of inputs, we vary the inputs during training.
For each sample, a random subset of the agents receives its context frames, possibly none, and the other views are
generated entirely.
At test time, the context frames of any subset of the agents can therefore be given, and for the agents without
context frames, the model generates their entire video.
Similarly, in training we drop the body pose of an agent at random and keep only its head motion, which we take
from its camera trajectory, so that the model can also generate agents for which no full body pose is available.
The joints of Body-Ray are dropped at random in the same way, so that the same model generates frames with and without
them.
When we scale the model with more data (\cref{sec:setup}), we also cut the links between views at random (link dropout), so that
the model can be run with only some of the views linked (details in the supplementary).
Since in most short sequences the agents rarely appear in each other's views, we oversample sequences in which an agent is
visible in another agent's view, drawing half of the training samples from them.

\section{Experiments}
\label{sec:experiments}
We evaluate \model against the three requirements described in \cref{sec:intro}: each agent follows its own controls, the motion
commanded for one agent appears coherently in the views of the others, and all views show a consistent world.

\subsection{Experimental setup}
\label{sec:setup}

\noindent
\textbf{Data.}
Training a multi-agent world model needs synchronized videos of several agents who act in the same scene, together with their
actions.
Such data is hard to record in the real world, which is why existing multi-agent models are mostly trained in games or
with fixed tabletop cameras~\cite{savva2026solaris, wu2026multiworld, liu2026gamma}.
Still, Nymeria~\cite{ma2024nymeria} provides such data for real people. All agents of a recording wear Aria
glasses~\cite{engel2023aria}, which are synchronized and localized in one world frame.
A portion of Nymeria has three agents, where two of them (A1, A2) perform daily tasks and a third (A3) follows and observes them.
The remaining larger part has two agents, A1 performing the tasks and A3 observing.
A1 and A2 also wear full-body motion capture suits, while A3 has no recorded body pose and acts more passively.
For \model, A3 is still an agent like the others: since training randomly drops the body pose of every agent, any agent can be controlled with or without its body pose, and A3 is controlled by its head motion only (\cref{sec:training}).
All models are trained on the three-agent recordings, with the train/test split of
EgoControl~\cite{pallotta2026egocontrol}. We use 300 three-agent clips from the test set for evaluations.
A scaled version of our model is additionally trained on the larger portion of Nymeria's two-agent recordings. We split these two-agent recordings based on the recording environments to additionally perform evaluation on unseen out-of-distribution environments.

\noindent
\textbf{Baselines.}
We compare \model with recent multi-agent video world models: Solaris~\cite{savva2026solaris},
MultiWorld~\cite{wu2026multiworld}, ShareVerse~\cite{zhu2026shareverse}, $\gamma$-World~\cite{liu2026gamma} and
Prisma-World~\cite{sun2026prisma}.
For a controlled comparison, we re-implement the multi-view mechanism of each method on our backbone and train it
with our data, inputs, and recipe.
We also report the base model EgoControl~\cite{pallotta2026egocontrol}, finetuned with the same training budget and applied to each view independently.

\noindent
\textbf{Implementation details.}
\model builds on EgoControl~\cite{pallotta2026egocontrol}, a diffusion transformer based on Cosmos-Predict2
2B~\cite{agarwal2025cosmos}.
Every view has $480\times480$ pixels at 16\,fps. The model is given 13 context frames and generates the following 32
frames of all views jointly.
The cross-view layers are initialized from the self-attention weights of EgoControl, with a zero-initialized output.
All models are trained for 40k iterations with a batch size of 16.
The scaled model keeps finetuning with Body-Ray for 40k more iterations, with a batch size of 64 and the two-agent
recordings added.
Sampling uses 35 steps without guidance, except where guided sampling is stated.
Further details are given in the supplementary material.

\begin{table*}[t]
\centering
\resizebox{\linewidth}{!}{%
\begin{tabular}{l|cc|cc|ccc|cc|cc}
\toprule
\multirow{3}{*}{\textbf{Method}} & \multicolumn{2}{c|}{\multirow{2}{*}{\textbf{Visual quality}}} & \multicolumn{2}{c|}{\multirow{2}{*}{\textbf{Camera control}}} & \multicolumn{3}{c|}{\multirow{2}{*}{\textbf{Action coherence, MPJPE (cm)}}} & \multicolumn{4}{c}{\textbf{Shared world consistency}} \\
 & \multicolumn{2}{c|}{} & \multicolumn{2}{c|}{} & \multicolumn{3}{c|}{} & \multicolumn{2}{c|}{All views with context} & \multicolumn{2}{c}{View transfer} \\
 & SSIM\up & FVD\down & TransErr\down & RotErr\down & No-edit\down & Single-edit\down & Multi-edit\down & Static (\%)\up & Dynamic (cm)\down & LPIPS\down & mIoU\up \\
\midrule
EgoControl~\cite{pallotta2026egocontrol} & 0.654 & 92 & 5.9 & 2.61 & 15.6 & 70.2 & 76.3 & 50.0 & 31.2 & 0.719 & 4.6 \\
MultiWorld~\cite{wu2026multiworld} & 0.658 & 89 & 5.7 & 2.51 & 15.5 & 70.5 & 76.0 & 50.0 & 31.1 & 0.708 & 5.8 \\
Solaris~\cite{savva2026solaris} & 0.659 & 78 & 5.9 & 2.58 & 11.1 & 59.1 & 70.9 & 51.0 & 28.5 & 0.646 & 8.6 \\
ShareVerse~\cite{zhu2026shareverse} & 0.680 & 73 & 5.7 & 2.06 & 9.9 & 50.4 & 60.9 & \underline{52.3} & 27.1 & 0.604 & 13.8 \\
$\gamma$-World~\cite{liu2026gamma} & 0.647 & 80 & 6.0 & 2.96 & 11.3 & 60.5 & 72.3 & 51.4 & 28.3 & 0.644 & 8.2 \\
Prisma-World~\cite{sun2026prisma} & 0.757 & 72 & \textbf{4.7} & 0.92 & 9.1 & 34.6 & 38.1 & 51.8 & 25.5 & 0.432 & 33.5 \\
\midrule
\model (ours) & \textbf{0.762} & \textbf{69} & \underline{5.2} & \textbf{0.90} & \underline{8.2} & \underline{30.2} & \underline{32.5} & \textbf{52.5} & \textbf{24.2} & \underline{0.387} & \underline{38.1} \\
\model + Body-Ray & \underline{0.761} & \underline{70} & 5.3 & \underline{0.91} & \textbf{8.0} & \textbf{29.0} & \textbf{32.2} & 51.6 & \underline{24.5} & \textbf{0.383} & \textbf{38.9} \\
\bottomrule
\end{tabular}}
\caption{\textbf{Comparison with multi-agent world models} on 300 three-agent test clips. All models are trained with the same
budget. Metrics are defined in \cref{sec:eval}.
}
\label{tab:main}
\end{table*}

\subsection{Evaluation Protocol}
\label{sec:eval}

We evaluate \model across several axes.

\noindent
\textbf{Visual quality and per-agent control.}
First, we evaluate whether multi-agent models preserve the visual quality and control of each agent's own body.
We measure this by calculating SSIM and FVD between generated and real ego videos.
Additionally, we measure how well the camera of each ego view, which is the head of its agent, follows the control trajectory (TransErr, RotErr)~\cite{he2025cameractrl}.
The rest of the body is only partly visible in its own view, so we measure it in the views of the other agents, as described next.

\begin{figure*}[t]
  \centering
  \includegraphics[width=\textwidth]{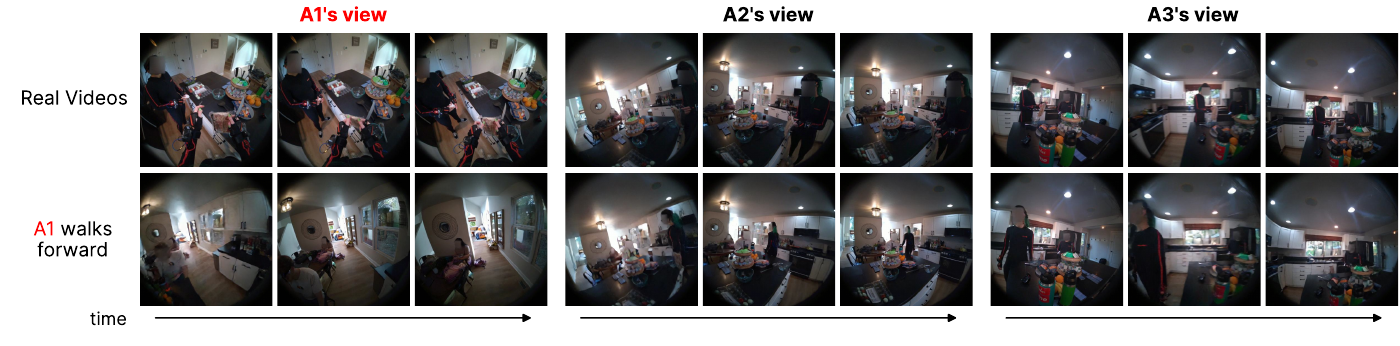}
  \caption{\textbf{Counterfactual edit.} Top: the real videos of a test clip. Bottom: our model generates the same clip from the same context frames, with the motion of \textbf{\textcolor{red}{A1 edited}} by walking
  forward. A1's view follows A1's new head motion, and A2 and A3 see A1 walk away, while in the real videos A1 stays at the counter. Three of the 32 generated frames are shown per view.}
  \label{fig:counterfactual}
\end{figure*}

\noindent
\textbf{Cross-agent action coherence.}
A multi-agent model must generate the same motion for an agent in their own view as for every other agent that sees it.
To measure this consistency, we run SAM 3D Body~\cite{yang2026sam} on each generated view, estimate the body of every other
agent visible in it, and compare it with a reference pose by the mean per-joint position error (MPJPE) over 12 body joints.
Based on this, we define three metrics: in \textbf{\textit{No-edit}}, every agent receives its recorded pose, and the reference is the pose estimated from the real
video.
This setting alone does not show that the model follows the controls, since an agent's motion can largely be
predicted from the context frames.
We therefore also evaluate with counterfactual edits (\cref{fig:counterfactual}).
In \textbf{\textit{Single-edit}} and \textbf{\textit{Multi-edit}}, the motion of one or more agents is replaced by a different
motion, which the context frames do not predict and which serves as the reference.
In Multi-edit, each edited agent must appear with its new motion in the view of the other, who is also
following a new motion.

\noindent
\textbf{Shared world consistency.}
All views must show the same world, both its static and dynamic parts.
For the static scene, we lift the pixels of one generated ego view to 3D with its estimated depth and camera, project them into the other ego view, and report the share of points whose depth differs by at most 20\% from the depth estimated there.
To measure consistency for dynamic parts, we estimate the pose of an agent in two generated views and measure the distance between them in world coordinates with a cross-view MPJPE.
To evaluate view transfer, we withhold the context frames of one ego view while providing context
to the other two views. We measure LPIPS and the semantic layout agreement (mIoU) against the
real target view. The mIoU is the mean IoU (\%) of the segmentation masks of 13 predefined classes, predicted by SAM3~\cite{carion2026sam}.
This setting tests whether the model can reconstruct an agent's surroundings from information shared by the other agents, without relying on its own visual history.
Further details of all metrics are given in the supplementary.

\subsection{Main Results}
\label{sec:comparison}

\Cref{tab:main} shows that \model reaches the best visual quality (SSIM 0.762, FVD 69) and is on par with the best baseline in camera control.
Considering action coherence, \model reproduces the poses of the other agents more accurately than all baselines in every
setting.
Its margin over the best baseline, Prisma-World~\cite{sun2026prisma}, widens as the setting gets harder, from
0.9\,cm without edits (8.2\,cm vs.\ 9.1\,cm) to 4.4\,cm after a single edit (30.2\,cm vs.\ 34.6\,cm) and 5.6\,cm
when both A1 and A2 are edited (32.5\,cm vs.\ 38.1\,cm).
Adding Body-Ray (\cref{sec:bodyray}) further improves
action coherence in every setting.
\Cref{fig:counterfactual} shows a qualitative example and we provide additional generated videos on our \href{https://cvg-bonn.github.io/MultiControl/}{project page}.

For the shared world consistency metrics, \model generates agents most consistently across views (Dynamic 24.2\,cm, against
25.5\,cm for Prisma-World).
This comes close to the error when calculating this metric on the ground-truth videos, 22.9\,cm, which comes from SAM 3D Body and camera estimates, showing that \model is highly consistent.
The static scene is about equally consistent for all methods (50.0\% to 52.5\%, \model the highest), since much of it is already visible in the context frames of each view.
In view transfer, where one ego view has no context frames of its own, \model improves over Prisma-World from 0.432 to 0.387 LPIPS and from 33.5\% to 38.1\% mIoU.

As we show in the next section, these gains come from our geometric position encoding, which relates the views at the level of individual tokens.
Both action coherence and view transfer rely on this fine-grained correspondence.

\subsection{Geometry Encoding Ablations}
\label{sec:ablations}

\begin{table}[t]
\centering
\resizebox{\linewidth}{!}{%
\setlength{\tabcolsep}{3pt}
\begin{tabular}{l|c|cc|cc|cc|cc}
\toprule
\multirow{2}{*}{\textbf{Geometry}} & \textbf{Quality} & \multicolumn{2}{c|}{\textbf{Camera control}} & \multicolumn{2}{c|}{\textbf{Action coherence}} & \multicolumn{2}{c|}{\textbf{Shared world}} & \multicolumn{2}{c}{\textbf{View transfer}} \\
 & SSIM\up & TransErr\down & RotErr\down & No-edit\down & Single\down & Static\up & Dynamic\down & LPIPS\down & mIoU\up \\
\midrule
ID embedding only & 0.661 & 5.7 & 2.53 & 9.4 & 44.8 & \underline{52.4} & 25.4 & 0.641 & 9.0 \\
Camera embedding & 0.670 & 5.8 & 2.12 & 8.9 & 31.8 & 52.0 & 24.8 & 0.550 & 19.4 \\
Pl\"ucker~\cite{he2025cameractrl} & 0.705 & 5.8 & 1.61 & 8.8 & 32.5 & \underline{52.4} & 24.9 & 0.540 & 21.7 \\
PRoPE~\cite{prope} & 0.743 & \textbf{5.2} & 0.94 & 8.6 & 30.5 & 51.9 & 25.1 & 0.477 & 28.3 \\
URoPE-4 & \underline{0.760} & \textbf{5.2} & \underline{0.92} & \underline{8.5} & 31.8 & \underline{52.4} & \underline{24.6} & \underline{0.398} & \underline{37.0} \\
URoPE-16 (joint attn) & 0.756 & \underline{5.3} & 1.08 & 9.6 & 37.5 & 50.7 & 27.3 & 0.406 & 36.0 \\
Body-Ray only & 0.695 & 5.8 & 1.73 & 8.6 & \textbf{29.2} & 51.2 & 26.8 & 0.684 & 7.4 \\
\midrule
URoPE-16 (cross-view) & \textbf{0.762} & \textbf{5.2} & \textbf{0.90} & \textbf{8.2} & \underline{30.2} & \textbf{52.5} & \textbf{24.2} & \textbf{0.387} & \textbf{38.1} \\
\bottomrule
\end{tabular}}
\caption{\textbf{Cross-view geometry encoding.} Only the way the camera geometry is given to the cross-view layer
changes, except in the joint-attention row and in the Body-Ray only row, which has no cross-view layer.
URoPE-4 and URoPE-16 lift each key to 4 and 16 depths (\cref{eq:lift}).
View transfer scores the agent view that receives no context frames, as in \cref{tab:main}, while all other columns use context for every view.}

\label{tab:camera_ablation}
\end{table}

We ablate in \cref{tab:camera_ablation} how the camera geometry is given to the cross-view layer, and includes the components of the main model.
With the ID embedding only, the other agents are poorly placed (44.8\,cm after a single edit).
Cameras as input features (camera embedding of \cref{eq:camcode}, Pl\"ucker rays) recover most of this, but camera
control and view transfer remain limited (RotErr $\ge$1.61$^\circ$, mIoU $\le$21.7\%).
Encoding the relative camera geometry directly in the attention improves both.
PRoPE, adopted by Prisma-World~\cite{sun2026prisma}, yields the first substantial improvement, and lifting each key to several depths (URoPE~\cite{xie2026urope}, \cref{eq:lift}) improves view transfer further, since token-level correspondence helps the target view recover the scene from the other agents.
With 16 depths, our model is best or tied on every column.
A joint self-attention over all views with the same encoding hurts action coherence (37.5\,cm).

\subsection{Further ablations}
\label{sec:final}

\begin{table}[t]
\centering
\resizebox{\linewidth}{!}{%
\setlength{\tabcolsep}{3pt}
\begin{tabular}{l|cc|ccc|cc}
\toprule
\textbf{Model} & FVD\down & RotErr\down & No-edit\down & Single\down & Multi\down & LPIPS\down & mIoU\up \\
\midrule
\multicolumn{8}{l}{\textbf{Three-agent test clips (places seen in training)}} \\
\model (ours) & 69 & 0.90 & 8.2 & 30.2 & 32.5 & 0.387 & 38.1 \\
+ Body-Ray & 70 & 0.91 & 8.0 & 29.0 & 32.2 & 0.383 & 38.9 \\
+ two-agent data (scaled) & \textbf{56} & 0.69 & 7.4 & \textbf{26.9} & \textbf{29.5} & 0.329 & 47.1 \\
+ CFG & 63 & \textbf{0.68} & \textbf{7.2} & 27.5 & 30.2 & \textbf{0.313} & \textbf{50.2} \\
\midrule
\multicolumn{8}{l}{\textbf{Two-agent test clips (places not seen in training)}} \\
\model (ours) & 127 & 1.88 & 6.7 & 21.2 & - & 0.525 & 22.5 \\
+ Body-Ray & 133 & 1.96 & 6.4 & 20.3 & - & 0.515 & 22.9 \\
+ two-agent data (scaled) & \textbf{103} & 1.29 & 5.2 & \textbf{18.9} & - & 0.446 & 33.2 \\
+ CFG & 105 & \textbf{1.27} & \textbf{4.9} & 19.1 & - & \textbf{0.424} & \textbf{36.9} \\
\bottomrule
\end{tabular}}
\caption{\textbf{Body-Ray, additional data and guidance.} Our model compared to further training with
two-agent recordings (scaled model), and with guided sampling.
Top: columns as in \cref{tab:main}.
Bottom: 300 two-agent clips (A1 and A3), where only A1 has a recorded body pose, so Multi does not apply.
FVD and RotErr of A1's view, No-edit and Single of A1 in A3's view, LPIPS and mIoU of A1's view generated from A3's
context only.}
\label{tab:scaling}
\end{table}

\Cref{tab:scaling} shows the effects of Body-Ray, training on additional data, and classifier-free guidance (CFG) on our model.
We find that Body-Ray gives a small gain in action coherence (single-edit error 30.2\,cm to 29.0\,cm).
Moreover, training on additional two-agent data brings the largest gain across all metrics.
The gain is especially notable on out-of-distribution environments, where FVD drops from 133 to 103, and mIoU rises from
22.9 to 33.2, while the single-edit error drops to 18.9\,cm.
CFG further improves view transfer (mIoU 47.1 to 50.2, and 33.2 to 36.9 on the unseen places).
Overall, scaling the data and compute makes the model more accurate and generalizable, with additional benefits from the other sources.

\subsection{Exo-to-Multi-Ego Generation}
\label{sec:flexible}

\begin{table}[t]
\centering
\resizebox{\linewidth}{!}{%
\begin{tabular}{l|cc|cc|ccc}
\toprule
\textbf{Setting} & \multicolumn{2}{c|}{\textbf{Fidelity}} & \multicolumn{2}{c|}{\textbf{Unseen by A3}} & \multicolumn{3}{c}{\textbf{MPJPE (cm)}} \\
 & PSNR\up & LPIPS\down & Surf.\%\up & Color\up & No-edit\down & Single\down & Multi\down \\
\midrule
Separate                             & 16.74 & 0.395 & 60.3 & 13.89 & 23.0 & 68.1 & 82.7 \\
Joint                                & \textbf{17.03} & \textbf{0.387} & \textbf{63.3} & \textbf{14.18} & \textbf{11.8} & \textbf{28.0} & \textbf{35.5} \\
\bottomrule
\end{tabular}}
\caption{\textbf{Exo-to-multi-ego generation.}
Fidelity: against the real video. Unseen by A3: depth agreement
(\%) and color PSNR (dB) between the views of A1 and A2 on surfaces A3 did not see. MPJPE: each of A1 and A2 in the
view of the other.}
\label{tab:flexible}
\end{table}

During training, a random subset of the agents receives its context frames (\cref{sec:training}).
As a result, our model can perform exo-to-multi-ego generation, where only A3, which observes A1 and A2 from outside,
is given its context frames, and the views of A1 and A2 are generated entirely.
The two views could also be generated one at a time, but independently generated views can contradict each other.
They have to agree on the parts of the scene that A3 did not see, and each agent has to appear in the view of the
other with the motion it performs.
\model generates them jointly, so they exchange this information through the cross-view attention.
\Cref{tab:flexible} compares joint generation with generating each view separately with A3.

For the scene, we take the surfaces that A1 and A2 see but A3 does not, and measure the share of points that the two
generated views place at the same depth and the PSNR of their colors.
Joint generation raises the depth agreement from 60.3\% to 63.3\%.
For the agents, we measure the pose error of each of A1 and A2 in the view of the other (\cref{sec:eval}).
A separately generated view of A1 has no access to the future motion of A2, and the single-edit error reaches
68.1\,cm.
Joint generation reduces it to 28.0\,cm.
It also improves the fidelity to the real video compared to the separate generation.

\section{Conclusion}
We presented \model, a multi-agent egocentric video world model in which every agent is controlled by its own 3D full-body pose.
Geometry-aware cross-view attention lets the views of all agents share one world, and a flexible training recipe lets the same model control agents with or without a body pose and generate the views of several agents from the context of another.
With a counterfactual-edit protocol, we show that the motion commanded for one agent appears coherently in the views of the others, more accurately than with recent multi-agent world models.
More broadly, \model is a first step toward simulating rich social interactions, in which every person perceives and reacts to the bodies and actions of those around them.
Future work can explore physical contact and interactions with objects in the multi-agent scenario.

{
    \small
    \bibliographystyle{ieeenat_fullname}
    \bibliography{main}

@article{hu2026multiplayer,
  title={Multiplayer interactive world models with representation autoencoders},
  author={Hu, Anthony and Volhejn, V{\'a}clav and Rahary, Adrien Ramanana and Mulder, Chris and Makkar, Aditya and Liao, Alyx and Royer, Am{\'e}lie and Orsini, Manu and Jelley, Adam and Alonso, Eloi and others},
  journal={arXiv preprint arXiv:2607.05352},
  year={2026}
}

@article{zhu2026shareverse,
  title={ShareVerse: Multi-Agent Consistent Video Generation for Shared World Modeling},
  author={Zhu, Jiayi and Zhang, Jianing and Yang, Yiying and Cheng, Wei and Yuan, Xiaoyun},
  journal={arXiv preprint arXiv:2603.02697},
  year={2026}
}

@article{liu2026gamma,
  title={Gamma-World: Generative Multi-Agent World Modeling Beyond Two Players},
  author={Liu, Fangfu and He, Kai and Shen, Tianchang and Cao, Tianshi and Fidler, Sanja and Duan, Yueqi and Gao, Jun and Gilitschenski, Igor and Wang, Zian and Ren, Xuanchi},
  journal={arXiv preprint arXiv:2605.28816},
  year={2026}
}

@article{sun2026prisma,
  title={Prisma-World: Camera-Controllable Multi-Agent Video World Model},
  author={Sun, Huiqiang and Peng, Zhan and Wu, Size and Wang, Kun and Liao, Kang and Wang, Dianyi and Zeng, Xingyu and Jin, Sheng and Li, Yangguang and Cao, Zhiguo and others},
  journal={arXiv preprint arXiv:2606.09507},
  year={2026}
}

@article{wu2026multiworld,
  title={Multiworld: Scalable multi-agent multi-view video world models},
  author={Wu, Haoyu and Yu, Jiwen and Zou, Yingtian and Liu, Xihui},
  journal={arXiv preprint arXiv:2604.18564},
  year={2026}
}

@article{savva2026solaris,
  title={Solaris: Building a multiplayer video world model in minecraft},
  author={Savva, Georgy and Michel, Oscar and Lu, Daohan and Waiwitlikhit, Suppakit and Meehan, Timothy and Mishra, Dhairya and Poddar, Srivats and Lu, Jack and Xie, Saining},
  journal={arXiv preprint arXiv:2602.22208},
  year={2026}
}

@article{hu2026metaworld,
  title={MetaWorld: Scaling Multi-Agent Video World Model from Single-view Video Data},
  author={Hu, Teng and Lu, Mingchun and Wang, Yating and Zhang, Jiangning and Hao, Jinkun and Pan, Ye and Yi, Ran and Ma, Lizhuang and Tao, Dacheng},
  journal={arXiv preprint arXiv:2606.02753},
  year={2026}
}

@inproceedings{pallotta2026egocontrol,
  title={Egocontrol: Controllable egocentric video generation via 3d full-body poses},
  author={Pallotta, Enrico and Azar, Sina Mokhtarzadeh and Doorenbos, Lars and Ozsoy, Serdar and Iqbal, Umar and Gall, Juergen},
  booktitle={Proceedings of the IEEE/CVF Conference on Computer Vision and Pattern Recognition},
  pages={4269--4279},
  year={2026}
}

@inproceedings{he2025cameractrl,
  title={Cameractrl: Enabling camera control for video diffusion models},
  author={He, Hao and Xu, Yinghao and Guo, Yuwei and Wetzstein, Gordon and Dai, Bo and Li, Hongsheng and Yang, Ceyuan},
  booktitle={International Conference on Learning Representations},
  volume={2025},
  pages={100433--100464},
  year={2025}
}

@inproceedings{wang2024motionctrl,
  title={Motionctrl: A unified and flexible motion controller for video generation},
  author={Wang, Zhouxia and Yuan, Ziyang and Wang, Xintao and Li, Yaowei and Chen, Tianshui and Xia, Menghan and Luo, Ping and Shan, Ying},
  booktitle={ACM SIGGRAPH 2024 Conference Papers},
  pages={1--11},
  year={2024}
}

@InProceedings{Kim_2026_CVPR,
    author    = {Kim, Byungjun and Kim, Taeksoo and Lee, Junyoung and Joo, Hanbyul},
    title     = {Dexterous World Models},
    booktitle = {Proceedings of the IEEE/CVF Conference on Computer Vision and Pattern Recognition (CVPR)},
    month     = {June},
    year      = {2026},
    pages     = {29663-29673}
}

@article{bai2026whole,
  title={Whole-body conditioned egocentric video prediction},
  author={Bai, Yutong and Tran, Danny and Bar, Amir and LeCun, Yann and Darrell, Trevor and Malik, Jitendra},
  journal={Advances in Neural Information Processing Systems},
  volume={38},
  pages={164375--164418},
  year={2026}
}

@article{tu2026playerone,
  title={Playerone: Egocentric world simulator},
  author={Tu, Yuanpeng and Luo, Hao and Chen, Xi and Bai, Xiang and Wang, Fan and Zhao, Hengshuang},
  journal={Advances in Neural Information Processing Systems},
  volume={38},
  pages={145235--145261},
  year={2026}
}

@inproceedings{grauman2022ego4d,
  title={Ego4d: Around the world in 3,000 hours of egocentric video},
  author={Grauman, Kristen and Westbury, Andrew and Byrne, Eugene and Chavis, Zachary and Furnari, Antonino and Girdhar, Rohit and Hamburger, Jackson and Jiang, Hao and Liu, Miao and Liu, Xingyu and others},
  booktitle={Proceedings of the IEEE/CVF conference on computer vision and pattern recognition},
  pages={18995--19012},
  year={2022}
}

@inproceedings{grauman2024ego,
  title={Ego-exo4d: Understanding skilled human activity from first-and third-person perspectives},
  author={Grauman, Kristen and Westbury, Andrew and Torresani, Lorenzo and Kitani, Kris and Malik, Jitendra and Afouras, Triantafyllos and Ashutosh, Kumar and Baiyya, Vijay and Bansal, Siddhant and Boote, Bikram and others},
  booktitle={Proceedings of the IEEE/CVF conference on computer vision and pattern recognition},
  pages={19383--19400},
  year={2024}
}

@InProceedings{Wang_2025_CVPR,
    author    = {Wang, Jianyuan and Chen, Minghao and Karaev, Nikita and Vedaldi, Andrea and Rupprecht, Christian and Novotny, David},
    title     = {VGGT: Visual Geometry Grounded Transformer},
    booktitle = {Proceedings of the IEEE/CVF Conference on Computer Vision and Pattern Recognition (CVPR)},
    month     = {June},
    year      = {2025},
    pages     = {5294-5306}
}

@article{agarwal2026cosmos,
  title={Cosmos 3: Omnimodal world models for physical ai},
  author={Agarwal, Niket and Ali, Arslan and Allen, Jon and Antolini, Martin and Aubame, Adeline and Azzolini, Alisson and Bai, Junjie and Bala, Maciej and Balaji, Yogesh and Bapst, Josh and others},
  journal={arXiv preprint arXiv:2606.02800},
  year={2026}
}

@inproceedings{li2026egocentric,
  title={Egocentric World Model for Photorealistic Hand Object Interaction Synthesis},
  author={Li, Dayou and Liu, Lulin and Liu, Bangya and Zhou, Shijie and Feng, Jiu and Lu, Ziqi and Zheng, Minghui and You, Chenyu and Fan, Zhiwen},
  booktitle={European Conference on Computer Vision},
  pages={527--544},
  year={2026},
  organization={Springer}
}

@inproceedings{bar2025navigation,
  title={Navigation world models},
  author={Bar, Amir and Zhou, Gaoyue and Tran, Danny and Darrell, Trevor and LeCun, Yann},
  booktitle={2025 IEEE/CVF Conference on Computer Vision and Pattern Recognition (CVPR)},
  pages={15791--15801},
  year={2025},
  organization={IEEE}
}

@misc{agarwal2025cosmos,
      title={Cosmos World Foundation Model Platform for Physical AI}, 
      author={NVIDIA and Niket Agarwal and Arslan Ali and Maciej Bala and Yogesh Balaji and Erik Barker and Tiffany Cai and Prithvijit Chattopadhyay and Yongxin Chen and Yin Cui and Yifan Ding and Daniel Dworakowski and Jiaojiao Fan and Michele Fenzi and Francesco Ferroni and Sanja Fidler and Dieter Fox and Songwei Ge and Yunhao Ge and Jinwei Gu and Siddharth Gururani and Ethan He and Jiahui Huang and Jacob Huffman and Pooya Jannaty and Jingyi Jin and Seung Wook Kim and Gergely Klár and Grace Lam and Shiyi Lan and Laura Leal-Taixe and Anqi Li and Zhaoshuo Li and Chen-Hsuan Lin and Tsung-Yi Lin and Huan Ling and Ming-Yu Liu and Xian Liu and Alice Luo and Qianli Ma and Hanzi Mao and Kaichun Mo and Arsalan Mousavian and Seungjun Nah and Sriharsha Niverty and David Page and Despoina Paschalidou and Zeeshan Patel and Lindsey Pavao and Morteza Ramezanali and Fitsum Reda and Xiaowei Ren and Vasanth Rao Naik Sabavat and Ed Schmerling and Stella Shi and Bartosz Stefaniak and Shitao Tang and Lyne Tchapmi and Przemek Tredak and Wei-Cheng Tseng and Jibin Varghese and Hao Wang and Haoxiang Wang and Heng Wang and Ting-Chun Wang and Fangyin Wei and Xinyue Wei and Jay Zhangjie Wu and Jiashu Xu and Wei Yang and Lin Yen-Chen and Xiaohui Zeng and Yu Zeng and Jing Zhang and Qinsheng Zhang and Yuxuan Zhang and Qingqing Zhao and Artur Zolkowski},
      year={2025},
      eprint={2501.03575},
      archivePrefix={arXiv},
      primaryClass={cs.CV},
      url={https://arxiv.org/abs/2501.03575}, 
}

@misc{engel2023aria,
      title={Project Aria: A New Tool for Egocentric Multi-Modal AI Research}, 
      author={Jakob Engel and Kiran Somasundaram and Michael Goesele and Albert Sun and Alexander Gamino and Andrew Turner and Arjang Talattof and Arnie Yuan and Bilal Souti and Brighid Meredith and Cheng Peng and Chris Sweeney and Cole Wilson and Dan Barnes and Daniel DeTone and David Caruso and Derek Valleroy and Dinesh Ginjupalli and Duncan Frost and Edward Miller and Elias Mueggler and Evgeniy Oleinik and Fan Zhang and Guruprasad Somasundaram and Gustavo Solaira and Harry Lanaras and Henry Howard-Jenkins and Huixuan Tang and Hyo Jin Kim and Jaime Rivera and Ji Luo and Jing Dong and Julian Straub and Kevin Bailey and Kevin Eckenhoff and Lingni Ma and Luis Pesqueira and Mark Schwesinger and Maurizio Monge and Nan Yang and Nick Charron and Nikhil Raina and Omkar Parkhi and Peter Borschowa and Pierre Moulon and Prince Gupta and Raul Mur-Artal and Robbie Pennington and Sachin Kulkarni and Sagar Miglani and Santosh Gondi and Saransh Solanki and Sean Diener and Shangyi Cheng and Simon Green and Steve Saarinen and Suvam Patra and Tassos Mourikis and Thomas Whelan and Tripti Singh and Vasileios Balntas and Vijay Baiyya and Wilson Dreewes and Xiaqing Pan and Yang Lou and Yipu Zhao and Yusuf Mansour and Yuyang Zou and Zhaoyang Lv and Zijian Wang and Mingfei Yan and Carl Ren and Renzo De Nardi and Richard Newcombe},
      year={2023},
      eprint={2308.13561},
      archivePrefix={arXiv},
      primaryClass={cs.HC},
      url={https://arxiv.org/abs/2308.13561}, 
}

@inproceedings{yang2026sam,
  title={Sam 3d body: Robust full-body human mesh recovery},
  author={Yang, Xitong and Kukreja, Devansh and Pinkus, Don and Fan, Taosha and Park, Jinhyung and Shin, Soyong and Cao, Jinkun and Liu, Jia-Wei and Ugrinovic, Nicol{\'a}s and Sagar, Anushka and others},
  booktitle={Proceedings of the IEEE/CVF Conference on Computer Vision and Pattern Recognition},
  pages={7209--7219},
  year={2026}
}

@inproceedings{carion2026sam,
  title={Sam 3: Segment anything with concepts},
  author={Carion, Nicolas and Gustafson, Laura and Hu, Yuan-Ting and Debnath, Shoubhik and Hu, Ronghang and Suris Coll-Vinent, Didac and Ryali, Chaitanya and Alwala, Kalyan Vasudev and Khedr, Haitham and Huang, Andrew and others},
  booktitle={International conference on learning representations},
  volume={2026},
  pages={138846--138923},
  year={2026}
}

@inbook{ma2024nymeria, title={Nymeria: A Massive Collection of Multimodal Egocentric Daily Motion in the Wild}, ISBN={9783031726910}, ISSN={1611-3349}, url={http://dx.doi.org/10.1007/978-3-031-72691-0_25}, DOI={10.1007/978-3-031-72691-0_25}, booktitle={Computer Vision – ECCV 2024}, publisher={Springer Nature Switzerland}, author={Ma, Lingni and Ye, Yuting and Hong, Fangzhou and Guzov, Vladimir and Jiang, Yifeng and Postyeni, Rowan and Pesqueira, Luis and Gamino, Alexander and Baiyya, Vijay and Kim, Hyo Jin and Bailey, Kevin and Fosas, David S. and Liu, C. Karen and Liu, Ziwei and Engel, Jakob and Nardi, Renzo De and Newcombe, Richard}, year={2024}, month=Nov, pages={445–465} }

@inproceedings{miyato2024gta,
title={{GTA}: A Geometry-Aware Attention Mechanism for Multi-View Transformers},
author={Takeru Miyato and Bernhard Jaeger and Max Welling and Andreas Geiger},
booktitle={The Twelfth International Conference on Learning Representations},
year={2024},
url={https://openreview.net/forum?id=uJVHygNeSZ}
}

@article{su2024roformer, title={RoFormer: Enhanced transformer with Rotary Position Embedding}, volume={568}, ISSN={0925-2312}, url={http://dx.doi.org/10.1016/j.neucom.2023.127063}, DOI={10.1016/j.neucom.2023.127063}, journal={Neurocomputing}, publisher={Elsevier BV}, author={Su, Jianlin and Ahmed, Murtadha and Lu, Yu and Pan, Shengfeng and Bo, Wen and Liu, Yunfeng}, year={2024}, month=Feb, pages={127063} }

@inproceedings{xie2026urope,
  title={Urope: Universal relative position embedding across geometric spaces},
  author={Xie, Yichen and Meng, Depu and Peng, Chensheng and Hu, Yihan and Herau, Quentin and Tomizuka, Masayoshi and Zhan, Wei},
  booktitle={European Conference on Computer Vision},
  pages={196--214},
  year={2026},
  organization={Springer}
}

@InProceedings{karras2024edm2,
    author    = {Karras, Tero and Aittala, Miika and Lehtinen, Jaakko and Hellsten, Janne and Aila, Timo and Laine, Samuli},
    title     = {Analyzing and Improving the Training Dynamics of Diffusion Models},
    booktitle = {Proceedings of the IEEE/CVF Conference on Computer Vision and Pattern Recognition (CVPR)},
    month     = {June},
    year      = {2024},
    pages     = {24174-24184}
}

@article{prope,
    title={Cameras as Relative Positional Encoding},
    author={Li, Ruilong and Yi, Brent and Liu, Junchen and Gao, Hang and Ma, Yi and Kanazawa, Angjoo},
    journal={Advances in Neural Information Processing Systems},
    year={2025}
}

@inproceedings{wu2026rayrope,
  title={Rayrope: Projective ray positional encoding for multi-view attention},
  author={Wu, Yu and Jeon, Minsik and Chang, Jen-Hao Rick and Tuzel, Oncel and Tulsiani, Shubham},
  booktitle={European Conference on Computer Vision},
  pages={192--210},
  year={2026},
  organization={Springer}
}

@InProceedings{Kong_2024_CVPR,
    author    = {Kong, Xin and Liu, Shikun and Lyu, Xiaoyang and Taher, Marwan and Qi, Xiaojuan and Davison, Andrew J.},
    title     = {{EscherNet}: A Generative Model for Scalable View Synthesis},
    booktitle = {Proceedings of the IEEE/CVF Conference on Computer Vision and Pattern Recognition (CVPR)},
    month     = {June},
    year      = {2024},
    pages     = {9503-9513}
}

@article{huang2025vipe,
  title={Vipe: Video pose engine for 3d geometric perception},
  author={Huang, Jiahui and Zhou, Qunjie and Rabeti, Hesam and Korovko, Aleksandr and Ling, Huan and Ren, Xuanchi and Shen, Tianchang and Gao, Jun and Slepichev, Dmitry and Lin, Chen-Hsuan and others},
  journal={arXiv preprint arXiv:2508.10934},
  year={2025}
}

@article{wang2026moge,
  title={Moge-2: Accurate monocular geometry with metric scale and sharp details},
  author={Wang, Ruicheng and Xu, Sicheng and Dong, Yue and Deng, Yu and Xiang, Jianfeng and Lv, Zelong and Sun, Guangzhong and Tong, Xin and Yang, Jiaolong},
  journal={Advances in Neural Information Processing Systems},
  volume={38},
  pages={35928--35959},
  year={2026}
}

@article{detone2026nymeriaplus,
  title={Nymeriaplus: Enriching nymeria dataset with additional annotations and data},
  author={DeTone, Daniel and Bogo, Federica and Le, Eric-Tuan and Frost, Duncan and Straub, Julian and Siddiqui, Yawar and Ye, Yuting and Engel, Jakob and Newcombe, Richard and Ma, Lingni},
  journal={arXiv preprint arXiv:2603.18496},
  year={2026}
}

@inproceedings{zhao2024rtdetr,
  title={Detrs beat yolos on real-time object detection},
  author={Zhao, Yian and Lv, Wenyu and Xu, Shangliang and Wei, Jinman and Wang, Guanzhong and Dang, Qingqing and Liu, Yi and Chen, Jie},
  booktitle={2024 IEEE/CVF Conference on Computer Vision and Pattern Recognition (CVPR)},
  pages={16965--16974},
  year={2024},
  organization={IEEE}
}

@article{xu2023vitpose,
  title={Vitpose++: Vision transformer for generic body pose estimation},
  author={Xu, Yufei and Zhang, Jing and Zhang, Qiming and Tao, Dacheng},
  journal={IEEE Transactions on Pattern Analysis and Machine Intelligence},
  volume={46},
  number={2},
  pages={1212--1230},
  year={2023},
  publisher={IEEE}
}

@article{khirodkar2024harmony4d,
  title={Harmony4d: A video dataset for in-the-wild close human interactions},
  author={Khirodkar, Rawal and Song, Jyun-Ting and Cao, Jinkun and Luo, Zhengyi and Kitani, Kris},
  journal={Advances in Neural Information Processing Systems},
  volume={37},
  pages={107270--107285},
  year={2024}
}

@article{huang2026self,
  title={Self forcing: Bridging the train-test gap in autoregressive video diffusion},
  author={Huang, Xun and Li, Zhengqi and He, Guande and Zhou, Mingyuan and Shechtman, Eli},
  journal={Advances in Neural Information Processing Systems},
  volume={38},
  pages={167283--167308},
  year={2026}
}
}

\clearpage
\setcounter{page}{1}
\maketitlesupplementary

\noindent
This supplementary material shows qualitative results
(\cref{sec:supp_qual}), reports additional results (\cref{sec:supp_results}), and gives implementation details of
\model{} and of the baselines (\cref{sec:supp_impl}) and the details of the evaluation (\cref{sec:supp_eval}).

\section{Qualitative results}
\label{sec:supp_qual}

\begin{figure*}[t]
\centering
\includegraphics[width=\linewidth]{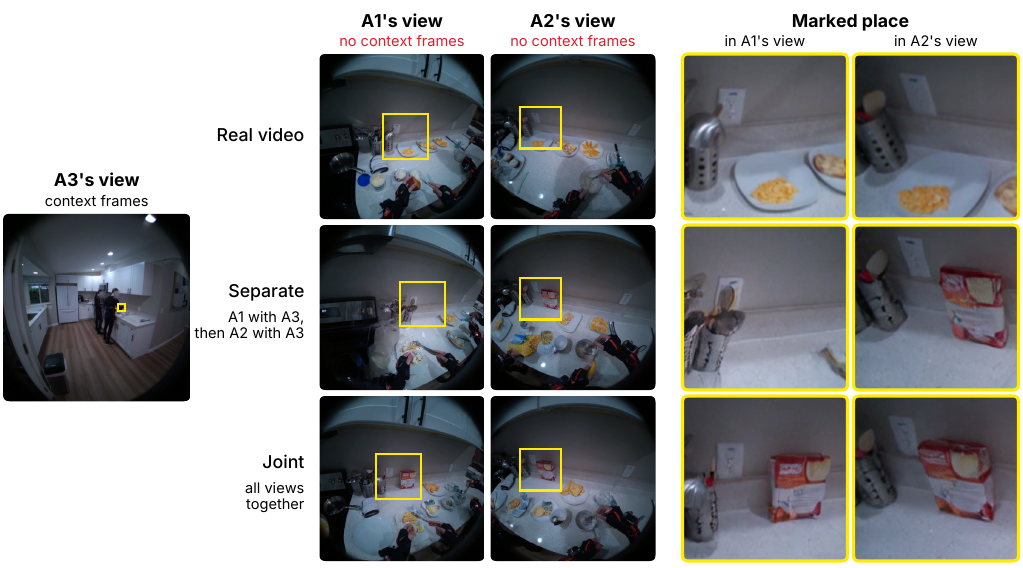}
\caption{\textbf{A place that A3 does not see, in exo-to-multi-ego generation.}
Only A3 receives its context frames.
The yellow boxes mark the place next to the utensil holders on the counter, and the crops on the right enlarge it.
In the view of A3, this place is hidden behind the agents.
In the real video, there are plates at this place.
With separate generation, the view of A2 shows a box there and the view of A1 does not.
With joint generation, both views show the same box.}
\label{fig:unseen}
\end{figure*}

\noindent
\textbf{Surfaces unseen by A3.}
In exo-to-multi-ego generation, the views of A1 and A2 contain surfaces that A3 does not see, and
\cref{tab:flexible} measures how well the two views agree on them.
\cref{fig:unseen} shows an example, generated by \model.
The context frames of A3 do not show the counter next to the utensil holders, so its content has to be invented.
With joint generation, the views of A1 and A2 show the same box there.
With separate generation, each of the two views is generated without the other, and only the view of A2 shows a box.

\begin{figure*}[t]
\centering
\includegraphics[width=\linewidth]{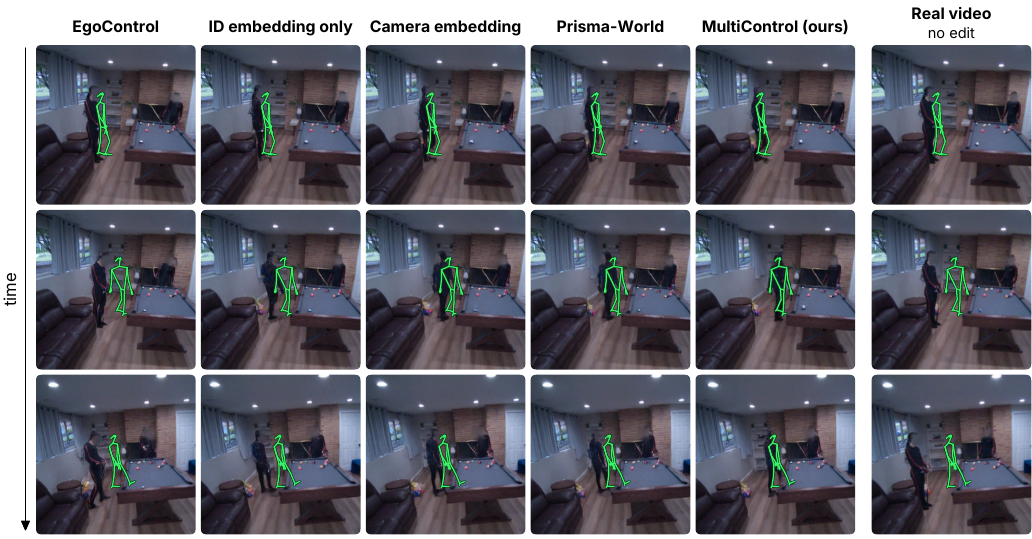}
\caption{\textbf{Counterfactual edit with different methods.}
The motion of A1 is edited so that A1 walks forward and to the left, and every method generates the clip from the same context frames.
The view of A3 is shown at frames 21, 33 and 45, cropped around A1.
The green skeleton is the edited pose of A1.
The real video shows the recorded motion of A1 without the edit.}
\label{fig:cmp_edit}
\end{figure*}

\begin{figure*}[t]
\centering
\includegraphics[width=\linewidth]{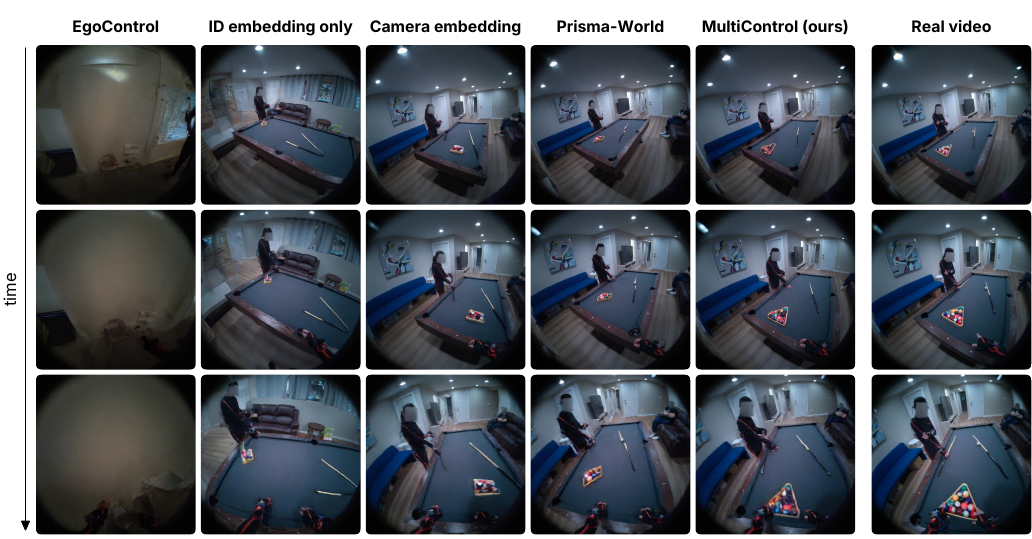}
\caption{\textbf{View transfer with different methods.}
The view of A2 is generated without its context frames, while A1 and A3 receive theirs.
Frames 21, 33 and 45 are shown.
Only \model places the rack with the balls on the table as in the real video.}
\label{fig:cmp_transfer}
\end{figure*}

\noindent
\textbf{Comparison with other approaches.}
\cref{fig:cmp_edit,fig:cmp_transfer} show qualitative comparisons of \model with EgoControl, ID embedding only, Camera embedding, and Prisma-World.
In \cref{fig:cmp_edit}, the motion of A1 is edited so that A1 walks forward and to the left.
With EgoControl and with the ID embedding only, A1 stays close to its place in the real video.
With the camera embedding and with Prisma-World, A1 walks but does not reach the edited pose, and with \model,
A1 is at the edited pose.
In \cref{fig:cmp_transfer}, the view of A2 is generated without its context frames.
EgoControl does not have access to other agent views and generates a different place. The ID embedding only shows the pool table from another side of the room.
The camera embedding and Prisma-World show the correct side of the room, but they place the rack with the balls at a
different position on the table, and \model places it as in the real video.

\noindent
\textbf{Video results.}
Our \href{https://cvg-bonn.github.io/MultiControl/}{project page} provides generated videos.
One video illustrates the method on a real sample.
The others show samples of \model for joint generation of all views, counterfactual edits,
exo-to-multi-ego generation with separate and joint generation, a full-body pose for A3, four and five agents at
test time, and the held-out places.
The videos also show samples on recordings of Harmony4D~\cite{khirodkar2024harmony4d}, which is not used for
training, and rollouts of 10.8\,s.
Finally, one video shows a sample of a variant of \model that we trained with Self Forcing~\cite{huang2026self} to
generate with 4 sampling steps.

\section{Additional results}
\label{sec:supp_results}

\noindent
\textbf{Links between views.}
Thanks to link dropout (\cref{sec:supp_impl}), the model can run with some links cut (\cref{tab:links}).
Without any link, each view is generated alone and the other agent does not follow its pose (Single-edit
64.5\,cm).
Linking A1 and A2 only through A3 is not enough (43.8\,cm) to flow information about A1 and A2 to each other and a direct link is needed.
With this direct link, also linking A3 leaves action coherence unchanged and slightly improves the ego views (FVD 58
to 56, RotErr 0.71$^\circ$ to 0.67$^\circ$).

\begin{table}[h!]
\centering
\resizebox{\linewidth}{!}{%
\setlength{\tabcolsep}{3pt}
\begin{tabular}{l|cc|c|ccc|c}
\toprule
\textbf{Links} & SSIM\up & FVD\down & RotErr\down & No-ed.\down & Single\down & Multi\down & Static\up \\
\midrule
None                                  & 0.621 & 84 & 4.12 & 13.4 & 64.5 & 82.5 & 48.9 \\
A1$\leftrightarrow$A2 only            & 0.783 & 58 & 0.71 & \textbf{6.9} & 27.8 & \textbf{34.8} & 52.1 \\
Via A3 only                           & 0.777 & 64 & 0.69 & 9.9 & 43.8 & 55.9 & 52.3 \\
All                           & \textbf{0.790} & \textbf{56} & \textbf{0.67} & 7.1 & \textbf{27.5} & \textbf{34.8} & \textbf{52.4} \\
\bottomrule
\end{tabular}}
\caption{\textbf{Links between views.}
None: no view is linked to another. A1$\leftrightarrow$A2 only: A3 is linked to no view. Via A3 only: A1 and A2 are
each linked to A3 but not to each other.
All metrics are computed on the ego views of A1 and A2, with the MPJPE of each of them in the view of the other.
Columns as in \cref{tab:main}. Bold: best.}
\label{tab:links}
\end{table}

\begin{table}[t]
\centering
\small
\setlength{\tabcolsep}{5pt}
\begin{tabular}{l|cc}
\toprule
\multirow{2}{*}{\textbf{Method}} & \multicolumn{2}{c}{\textbf{View transfer}} \\
 & LPIPS\down & mIoU\up \\
\midrule
\multicolumn{3}{l}{\textit{Methods of \cref{tab:main}}} \\
\cmidrule{1-3}
EgoControl~\cite{pallotta2026egocontrol}     & 0.719 & 4.6 \\
MultiWorld~\cite{wu2026multiworld}           & 0.707 & 5.9 \\
Solaris~\cite{savva2026solaris}              & 0.659 & 6.8 \\
ShareVerse~\cite{zhu2026shareverse}          & 0.627 & 10.9 \\
$\gamma$-World~\cite{liu2026gamma}           & 0.658 & 6.4 \\
Prisma-World~\cite{sun2026prisma}            & 0.480 & 27.7 \\
\model (ours)                                & \underline{0.453} & \underline{31.5} \\
\model + Body-Ray                            & \textbf{0.448} & \textbf{31.8} \\
\midrule
\multicolumn{3}{l}{\textit{Other geometry encodings of \cref{tab:camera_ablation}}} \\
\cmidrule{1-3}
ID embedding only                            & 0.661 & 7.0 \\
Camera embedding                             & 0.579 & 16.1 \\
Pl\"ucker~\cite{he2025cameractrl}            & 0.574 & 18.3 \\
PRoPE~\cite{prope}                           & 0.516 & 25.2 \\
URoPE-4                                      & 0.465 & 30.6 \\
URoPE-16 (joint attn)                        & 0.460 & 30.2 \\
Body-Ray only                                & 0.684 & 7.5 \\
\bottomrule
\end{tabular}
\caption{\textbf{Exo-to-multi-ego} for the methods of \cref{tab:main} and the geometry encodings of
\cref{tab:camera_ablation}, on the 300 three-agent test clips.
Only A3 is given its context frames, and the views of A1 and A2 are generated entirely.
LPIPS and mIoU (13 room classes, \%) compare the generated ego views with the real videos.
Bold: best. Underlined: second best.}
\label{tab:exo_baselines}
\end{table}

\noindent
\textbf{Exo-to-multi-ego for all methods.}
\Cref{tab:exo_baselines} compares the methods of \cref{tab:main} when the views of A1 and A2 are generated from the
context frames of A3 only.
The ranking is the same as in the main results, and \model improves over Prisma-World from 0.480 to
0.453 LPIPS and from 27.7\% to 31.5\% mIoU.
Among the geometry encodings, both metrics improve from the ID embedding only (0.661 LPIPS, 7.0\% mIoU) to PRoPE
(0.516, 25.2\%) and further to URoPE (0.453, 31.5\% with 16 depths).

\begin{table*}[t]
\centering
\resizebox{\linewidth}{!}{%
\begin{tabular}{l|cc|cc|ccc|cc|cc}
\toprule
\multirow{3}{*}{\textbf{Cross-view layer}} & \multicolumn{2}{c|}{\multirow{2}{*}{\textbf{Visual quality}}} & \multicolumn{2}{c|}{\multirow{2}{*}{\textbf{Camera control}}} & \multicolumn{3}{c|}{\multirow{2}{*}{\textbf{Action coherence, MPJPE (cm)}}} & \multicolumn{4}{c}{\textbf{Shared world consistency}} \\
 & \multicolumn{2}{c|}{} & \multicolumn{2}{c|}{} & \multicolumn{3}{c|}{} & \multicolumn{2}{c|}{All views with context} & \multicolumn{2}{c}{View transfer} \\
 & SSIM\up & FVD\down & TransErr\down & RotErr\down & No-edit\down & Single-edit\down & Multi-edit\down & Static (\%)\up & Dynamic (cm)\down & LPIPS\down & mIoU\up \\
\midrule
ID only                                                                  & 0.653 & \underline{78} & 6.0 & 2.76 & 10.2 & 52.0 & 63.6 & 51.0 & 26.7 & 0.652 & 8.2 \\
ID + camera embedding, $\phi=(\mathbf{c},\log(1{+}\lVert\mathbf{c}\rVert),R^{(1)},R^{(2)})$ & \textbf{0.657} & \underline{78} & \underline{5.9} & 2.67 & \underline{9.4} & \textbf{36.0} & \textbf{40.4} & 51.4 & \textbf{24.7} & 0.624 & \underline{11.9} \\
ID + camera embedding, $\phi=(\mathbf{c},R^{(1)},R^{(2)})$               & 0.655 & \textbf{77} & \underline{5.9} & \textbf{2.62} & 9.5 & \underline{38.0} & \underline{41.4} & \underline{51.9} & \underline{25.5} & \textbf{0.620} & 11.7 \\
ID + camera embedding, $\phi=(\mathbf{c},\log(1{+}\lVert\mathbf{c}\rVert),R)$ & 0.655 & \textbf{77} & \underline{5.9} & 2.76 & \underline{9.4} & 38.9 & 42.6 & \underline{51.9} & 25.6 & \underline{0.623} & 10.9 \\
ID + camera embedding, $\phi=(\mathbf{c},R)$                             & \underline{0.656} & \textbf{77} & \textbf{5.7} & \underline{2.64} & \textbf{9.2} & 38.7 & 42.6 & \textbf{52.1} & 25.8 & \underline{0.623} & \textbf{12.1} \\
\bottomrule
\end{tabular}}
\caption{\textbf{Camera embedding} after 20k iterations, on the 300 three-agent test clips. Every model is trained with the same recipe and budget, and only the listed parts change. The cross-view layer
uses the 3D RoPE of the base model at each token's own position instead of URoPE, with the ID embedding alone or with
camera embeddings computed from different inputs $\phi$ (\cref{eq:camcode}), where $R^{(1)}$ and $R^{(2)}$ are the
first two columns of $R$. Columns as in \cref{tab:main}. Bold: best. Underlined: second best.}
\label{tab:abl20k}
\end{table*}

\noindent
\textbf{Camera embedding.}
\Cref{tab:abl20k} compares cross-view layers that use the 3D RoPE of the base model instead of URoPE, each trained for
20k iterations.
The camera embedding mainly improves action coherence.
With the ID embedding alone, the Single-edit error is 52.0\,cm and the Multi-edit error 63.6\,cm, and the camera
embedding reduces them to 36.0\,cm and 40.4\,cm.
The input $\phi$ of \cref{eq:camcode} (row 2) gives the lowest errors.
The three other inputs reach 38.0\,cm to 38.9\,cm for Single-edit and 41.4\,cm to 42.6\,cm for Multi-edit.

\noindent
\textbf{Single-edit error per motion.}
\Cref{tab:permotion} splits the Single-edit error of \cref{tab:main} by the motion of the edit.
\model is more accurate than Prisma-World for five of the seven motions, most of all for walking forward (29.8\,cm
against 42.1\,cm) and walking forward and to the left (27.2\,cm against 37.8\,cm).
It is less accurate for sitting down (41.8\,cm against 41.2\,cm) and for arm motion (30.7\,cm against 27.2\,cm).
With Body-Ray, the error for arm motion drops to 27.5\,cm.

\begin{table}[t]
\centering
\resizebox{\linewidth}{!}{%
\setlength{\tabcolsep}{3pt}
\begin{tabular}{l|ccccccc}
\toprule
\textbf{Method} & Walk forward & Walk left & Turn & Sit & Crouch & Arms & Look \\
\midrule
EgoControl             & 130.3 & 97.3 & 47.4 & 68.1 & 49.8 & 39.3 & 39.5 \\
MultiWorld             & 129.5 & 99.3 & 47.3 & 67.5 & 51.1 & 39.8 & 39.0 \\
Solaris                & 105.2 & 85.5 & 42.0 & 61.0 & 50.9 & 39.2 & 34.4 \\
ShareVerse             & 79.4 & 64.4 & 35.8 & 55.9 & 46.4 & 37.9 & 28.6 \\
$\gamma$-World         & 114.7 & 90.9 & 42.5 & 62.4 & 48.3 & 40.2 & 36.2 \\
Prisma-World           & 42.1 & 37.8 & 26.0 & \textbf{41.2} & 40.5 & \textbf{27.2} & \underline{21.7} \\
\model (ours)          & \underline{29.8} & \textbf{27.2} & \underline{21.7} & 41.8 & \underline{36.0} & 30.7 & \textbf{17.3} \\
\model + Body-Ray      & \textbf{28.9} & \underline{27.6} & \textbf{21.1} & \underline{41.7} & \textbf{35.7} & \underline{27.5} & \textbf{17.3} \\
\bottomrule
\end{tabular}}
\caption{\textbf{Single-edit error per motion.} The Single-edit MPJPE of \cref{tab:main} (cm, lower is better),
split by the motion that the edit introduces.
Bold: best. Underlined: second best.}
\label{tab:permotion}
\end{table}

\section{Implementation details}
\label{sec:supp_impl}

\noindent
\textbf{Backbone.}
\model{} builds on EgoControl~\cite{pallotta2026egocontrol}, which is based on Cosmos-Predict2
2B~\cite{agarwal2025cosmos}, a diffusion transformer with 28 blocks and 16 attention heads of dimension $D=128$.
The video autoencoder of the base model compresses a video by a factor of 4 in time and 8 in space, and the
transformer works on patches of $2\times2$ latent cells.

\noindent
\textbf{Videos and tokens.}
Every view is generated at $480\times480$ pixels and 16\,fps.
A clip has 45 frames, 13 context frames followed by 32 frames to generate.
They become 12 latent frames, the first 4 of which hold the context, and each latent frame gives a grid of
$30\times 30$ tokens.

\noindent
\textbf{Cross-view attention.}
Each block has a pool of 8 ID embeddings.
In training, the views of a sample draw different IDs at random, and at test time view $i$ uses the $i$-th.
The ID embeddings and the output layer of the camera MLP $\psi$ start at zero, and $\psi$ has one hidden layer
of width 256.
Each cross-view layer is initialized from the self-attention layer of its block in the pretrained model, except for
its output projection $W^{\mathrm{G}}_o$ (\cref{eq:crossview}), which is initialized with zeros, so that training
starts from the base model.
The 16 heads of the cross-view layer assume the depths $d_h=0.25\cdot64^{h/15}$\,m, $h=0,\dots,15$, from 0.25\,m to
16\,m.
Each latent frame has one camera pose, taken at the center of the video frames it holds.
Points behind the query camera are projected with their true angle from the forward axis, capped at 1.4\,rad,
and projected positions are clamped to $[-2\times 30-\tfrac12,\ 3\times 30-\tfrac12]$ along each axis.
The layer accepts any number of views, since a view attends to the tokens of all other views in one softmax
(\cref{eq:crossview}).%

\noindent
\textbf{Body-Ray.}
The joint positions of \cref{eq:bodykey} are sampled like for camera poses, one per latent frame.
The rotation $\mathcal{R}$ of \cref{eq:bodyray} uses 32 frequencies per angle, $\omega_\kappa=2\cdot16^{\kappa/31}$,
$\kappa=0,\dots,31$.
Body-Ray follows the links between views, so view $i$ receives the joints of agent $j$ only if it attends to view $j$.

\noindent
\textbf{Training data.}
The three-agent training set has 28,983 windows from 84 pairs of recordings.
The windows that are oversampled (\cref{sec:training}) are those in which A1 or A2 is visible in at least half of the
frames of the other's view, taking the view in which this agent is most prominent, and moves there by at least 2 pixels per frame in the 480-pixel image while visible.
An agent counts as visible in a frame when at least $1/4$ of its joints project into the image.
Its motion is computed from its joints at two consecutive frames, both projected with the camera of the first frame,
so that the head motion of the viewing agent does not count.
This filter selects 7,668 windows with strong multi-agent dynamics.
Half of the three-agent samples are drawn uniformly from this set, and the other half uniformly from the remaining 21,315
windows.
For the scaled model (\cref{sec:setup}), 70\% of the training iterations use two-agent samples and the other 30\%
use three-agent samples, drawn as above.
The two-agent samples come from all two-agent recordings, excluding those at the four held-out places that we keep for testing.
This gives 268,625 windows from 808 recordings at 38 places.

\noindent
\textbf{Flexible inputs in training.}
For every training sample, we choose at random which views receive their context frames.
For $N=3$ views, each of the eight possible subsets ($2^N$) of views is equally likely, including the one in which no view receives context frames.
A view without context frames is handled in one of two ways, each with probability 0.5.
Either all its latent frames are generated and scored, or its context latents are replaced by zeros that are still
marked as given (the context dropout of the base model~\cite{pallotta2026egocontrol}).
Each view draws its own noise level.
An agent with a body pose is given only the head motion with probability 0.1.
It then receives the same input as A3, which has no recorded body pose, so only the head motion remains.
The head entry of every agent is computed from its camera trajectory, rotated by one fixed rotation into the axes of
the motion-capture head.
The Body-Ray update of a view is set to zero with probability 0.5, so that the same model generates with and without the joints.

\noindent
\textbf{Link dropout.}
In the training of the scaled model, the links between views are cut at random, so that the model can be run with
only some of the views linked (\cref{tab:links}).
When the link between two views is cut, neither view attends to the other, and neither receives the joints of the
other agent through Body-Ray.
For a three-view sample, all links are kept with probability 0.5.
Otherwise, one of five patterns is drawn, each with probability 0.1.
The patterns are no links, only A1 and A3 linked, only A2 and A3 linked, only A1 and A2 linked, and A3 linked to both
A1 and A2 with the link between A1 and A2 cut.
For a two-view sample, the link is cut with probability 0.1.
A view without any link receives no cross-view update.

\noindent
\textbf{Optimization and training schedule.}
All models start from the base model's 480p weights and train all parameters with AdamW ($\beta_1=0.9$,
$\beta_2=0.99$, weight decay 0.1), gradient clipping at 1.0 and bfloat16 precision.
The new layers use a learning rate of $10^{-4}$, and the pretrained weights use $7\times10^{-6}$, with a linear
warm-up over 500 iterations and a linear decay to 0.1 of the peak over each block of 20k iterations.
We keep a power-function exponential moving average of the weights~\cite{karras2024edm2} with
$\sigma_{\mathrm{rel}}=0.1$ and evaluate the averaged weights.
All models are trained for 40k iterations with one sample of three views per GPU on 16 GPUs (global batch size 16).
The Body-Ray version starts from our model after 20k iterations and trains for 20k more with Body-Ray branch enabled.
The scaled model continues it for 40k iterations on 64 GPUs (batch size 64), with the two-agent recordings
and link dropout, and a learning rate of $3\times10^{-5}$ for the pretrained weights that decays to 0.3 of the peak.

\noindent
\textbf{Baselines.}
All baselines of \cref{tab:main} start from the same EgoControl weights as \model{} and use the same data, three
views per sample, flexible inputs, optimizer and number of iterations, and all their parameters are trained.
Every agent is controlled through the pose conditioning of EgoControl, with A3 given by its head motion.
Only the mechanism that couples the views is taken from the respective method.
\textit{EgoControl}~\cite{pallotta2026egocontrol} has no such mechanism, and each view is generated independently.
\textit{Solaris}~\cite{savva2026solaris} extends the self-attention of every block over the tokens of all views, with the same position encoding for each view, and adds a learned embedding per agent to the tokens at the input of every block.
\textit{$\gamma$-World}~\cite{liu2026gamma} also uses one self-attention over all views.
It encodes the agent in the position encoding, where the low temporal frequencies of RoPE are replaced by a code of
the agent, the vertex of a simplex.
We use the dense attention of its bidirectional model.
\textit{ShareVerse}~\cite{zhu2026shareverse} keeps the blocks of each view separate.
Before every second block, it adds an embedding of a ray map, which holds the ray direction and the camera center of
every pixel in world coordinates, to the tokens, followed by a joint attention over the tokens of all views, concatenated along the time axis.
\textit{Prisma-World}~\cite{sun2026prisma} uses a self-attention over all views in which the position grid of each agent is shifted, and adds a parallel attention branch whose queries and keys are multiplied by the camera extrinsics.
\textit{MultiWorld}~\cite{wu2026multiworld} has no attention between views.
The poses of all agents are encoded and combined across agents into one token per frame, which every view reads through a cross-attention in each block.
Its global state encoder is a frozen VGGT~\cite{Wang_2025_CVPR} applied to the first context frame of every view that
receives context frames, and its tokens are read through a second cross-attention in each block.

\noindent
\textbf{Geometry encodings.}
The first five rows of \cref{tab:camera_ablation} and our model use the cross-view layer of \cref{sec:crossview} with
its ID embedding, and differ in how the camera geometry enters this layer.
All of them except the first row include the camera embedding $\mu_{ij}$ (\cref{eq:camcode}).
In the first three rows, queries and keys are rotated by the 3D RoPE of the base model at their own positions.
The Pl\"ucker row also adds an embedding of the Pl\"ucker ray of every token, expressed in the coordinates of the
query view, to the queries, keys and values.
PRoPE multiplies the queries and keys by the camera matrices of their views in 60 of the 128 channels of each head
and keeps the 3D RoPE in the other channels.
URoPE-4 uses the depths 0.75, 1.5, 3 and 6\,m, each for four heads.
In the joint-attention row, the self-attention of the base model spans the tokens of all views, with the ID and
camera embeddings and with URoPE-16 between tokens of different views without a separate cross-view layer.
The Body-Ray only row has no attention between views, and the views are coupled through Body-Ray alone.

\section{Evaluation details}
\label{sec:supp_eval}

We now describe all the metrics used to evaluate our method in detail.
Unless stated otherwise, every view is given its 13 context frames, and the metrics are computed on the 32 generated frames of the 300 three-agent test clips.
Camera paths, depth, and body poses are estimated in the same way on the generated and on the real videos, after the fisheye frames are remapped to a pinhole image with the calibrated camera model.

\noindent
\textbf{Test clips.}
The three-agent test clips are 300 windows of the test split of EgoControl~\cite{pallotta2026egocontrol},
chosen among the windows in which A1 and A2 see each other in at least half of the frames.
The two-agent test clips of \cref{tab:scaling} are 300 windows from 45 recordings at four places that are held out from training, 75 per place.

\noindent
\textbf{Camera control.}
The control trajectory of a view is the recorded camera path of its real video.
We therefore estimate the camera path of the generated and of the real video with ViPE~\cite{huang2025vipe} from the
45 frames of the view, given the known intrinsics.
The generated path is aligned to the real one by a rigid transform that is fitted on the context frames, and its scale
is kept.
TransErr (cm) and RotErr (degrees) are the mean position and rotation errors between the two paths over the generated
frames, averaged over the ego views of all clips.

\noindent
\textbf{Counterfactual edits.}
An edit replaces the motion of an agent during the generated frames by one of seven motions, which are taken from
real recordings outside the test clips.
The motions are walking forward, walking forward and to the left, turning to the right on the spot, sitting down, crouching
and standing up again, moving both arms, and looking around while standing.
The joint rotations of the new motion are applied to the skeleton of the edited agent, so the agent keeps its own
body proportions.
The motion is blended in over at most the first 16 generated frames, and the camera trajectory of the edited agent
follows its new head motion.
An edit therefore changes $P_i$ and $T_i$ of the edited agent, while the context frames and the controls of the other
agents stay the same.
Each test clip is generated in seven versions, one without an edit, two that edit A1, two that edit A2, and two that
edit both, each with a different motion.
The assignment of motions to clips is drawn once, is balanced over the seven motions, and is the same for all methods.

\noindent
\textbf{Action coherence.}
A multi-agent model must generate the same motion for an agent in their own view as for every other agent that sees it. To measure this consistency, we first detect people with RT-DETR~\cite{zhao2024rtdetr} and ViTPose++~\cite{xu2023vitpose}. Then, each detection is
assigned to an agent by matching its 2D keypoints with the projection of the agent's reference pose into the view.
SAM 3D Body~\cite{yang2026sam} then estimates the 3D joints of every detected person in the coordinates of the camera.
We use 12 of these joints: the shoulders, elbows, wrists, hips, knees, and ankles, on every fourth generated frame, and measure the MPJPE, which is the mean distance between the estimated and the reference joints.
We report the median MPJPE over all person-frames in which the agent is found, in cm, with each of A1 and A2 evaluated in the view of the other one and also from the view of A3.

In No-edit, the reference is the pose estimated on the real video at the same view and frame.
The pose estimates of the real and of the generated video are both placed in the room with the estimated camera of
their video, after the camera path of the generated video is aligned to that of the real video.

In Single-edit and Multi-edit, the reference is the edited pose of the agent in the control camera of the viewing agent.
In Multi-edit, the viewing agent follows a new motion as well, and its control camera is the edited one.
The reference joints are taken from the body fits of NymeriaPlus~\cite{detone2026nymeriaplus}.
These fits use the body model that SAM 3D Body predicts, so estimated and reference joints have the same definition.
We evaluate frames 33, 37, 41 and 45, when the new motion is fully blended in, and only person-frames in which the
edited and the original pose are both visible in the view and at least 0.2 body sizes apart in the image, so that
following the edit can be told apart from ignoring it.
This gives 5,690 person-frames for Single-edit and 4,214 for Multi-edit.
On the real videos, in which the agents follow their recorded poses, the MPJPE to the recorded pose on the same
person-frames is 24.7\,cm for Single-edit and 25.4\,cm for Multi-edit.
This is the error of the estimator itself.
Since the bodies are not aligned, it includes the error in the estimated distance of the person from the camera,
which is ambiguous in a single image.

\noindent
\textbf{Shared-world consistency.}
All views must show the same consistent world, in both its static and dynamic parts.
For the static part, we estimate the depth of each view at frames 21, 33, and 45 with MoGe-2~\cite{wang2026moge} on
a grid of $120\times120$ cells.
The scale of a monocular depth estimate is not reliable, so the depth of a generated view is
rescaled per frame to the median depth of the real video of that view.
Each cell with a valid depth is lifted to 3D with the camera of its view and projected into the other view.
The camera path of a view is estimated from its video and aligned to its control trajectory as for camera control.
The \textit{Static} metric measures the share of the cells landing inside the other view at which the depth of the other view agrees within 20\%, pooled over both directions, the three frames and all clips. A1 and A2 views are used to calculate this metric.
On the real videos, \textit{Static} is 57.2\%, and it is 24.5\% when the second view is taken from another moment of the same
recording.

For the dynamic part, the 12 joints of an agent are estimated in two generated views that see it, on the videos
without edits, and each estimate is placed in the world with the estimated camera of its view.
\textit{Dynamic} measures the mean distance between the joints of the two estimates, as the median over person-frames, in cm.
A1 and A2 are compared between the view of the other and the view of A3, and the body of A3 between the views of A1
and A2.
The metric compares generated views with each other and needs no ground truth.
On the real videos it is 22.9\,cm, which is the error of the estimators.

\noindent
\textbf{View transfer.}
In this setting, one ego view receives no context frames, while the other two views do receive theirs.
Then, all 45 frames of the ego view without context are generated, and we evaluate frames 14 to 45 as in the other settings.
We report the LPIPS, which compares the generated frames with the real frames inside the image circle of the fisheye lens.
Additionally, we calculate the mIoU by using SAM~3~\cite{carion2026sam} to segment 13 room classes (floor, wall, ceiling, door, window, table, chair, sofa, bed, cabinet, shelf, countertop, and television) in the generated and in the real frames at frame 21, 33, and 45.
We report the IoU of the masks in \%, as the mean over the classes present in either frame.
Both metrics are averaged over the two choices of the ego view without context frames.

\noindent
\textbf{Exo-to-multi-ego.}
In this scenario, only A3 receives its context frames, and the views of A1 and A2 are generated entirely.
For the fidelity, we compare frames 14 to 45 of the two generated ego views with the real videos and report PSNR and LPIPS.

\cref{tab:flexible} compares two settings.
In \textit{Joint}, the views of A1, A2 and A3 are generated together, so A1 and A2 exchange information through the cross-view attention.
In \textit{Separate}, the view of A1 is generated together with the view of A3 only, and the view of A2 in a second run with the view of A3 only, so A1 and A2 exchange no information.
The difference between the two settings shows what this exchange adds.
We measure it where A3 cannot help, on the surfaces that A3 did not see and on the motion of the other agent.

For the surfaces, we take those that A1 and A2 both see but A3 does not.
We find them in the real videos, with the depth, the grid of cells and the frames of the \textit{Static} metric.
Cells that show a person are left out.
A cell of one ego view is seen by the other ego view if its 3D point, projected into that view, agrees with the depth there within 20\%.
It is unseen by A3 if the same test fails in the view of A3 at a context frame.
This holds for 37\% of the cells that both ego views see.
On these cells, we report two numbers for the two generated ego views.
The first is the share of cells at which their depths agree within 20\%, computed in the same way as the \textit{Static} metric.
The second is the PSNR between their colors.

For the motion, the three MPJPE columns of \cref{tab:flexible} follow the action coherence evaluation described above.
Here, each of A1 and A2 is evaluated only in the view of the other, and not in the view of A3.

\noindent
\textbf{Two-agent test clips.}
In the two-agent test clips, only A1 has a recorded body pose.
FVD and RotErr are computed on the view of A1, No-edit and Single-edit for A1 in the view of A3, and LPIPS and mIoU on the view of A1 generated from the context frames of A3 only.

\end{document}